\documentclass[letterpaper]{article} % DO NOT CHANGE THIS
\usepackage{aaai24}  % DO NOT CHANGE THIS
\usepackage{times}  % DO NOT CHANGE THIS
\usepackage{helvet}  % DO NOT CHANGE THIS
\usepackage{courier}  % DO NOT CHANGE THIS
\usepackage[hyphens]{url}  % DO NOT CHANGE THIS
\usepackage{graphicx} % DO NOT CHANGE THIS
\usepackage{natbib}  % DO NOT CHANGE THIS AND DO NOT ADD ANY OPTIONS TO IT
\usepackage{caption} % DO NOT CHANGE THIS AND DO NOT ADD ANY OPTIONS TO IT
\usepackage{algorithm}
\usepackage{algorithmic}

\usepackage{amsmath}
\usepackage{amssymb}
\usepackage{multicol, multirow}
\usepackage{booktabs}
\usepackage{colortbl}
\definecolor{Gray}{gray}{0.9}
\usepackage{enumitem}
\usepackage{nccmath}
\usepackage{textcomp}
\usepackage{subfigure}
\usepackage[dvipsnames]{xcolor}

\definecolor{myyellow}{RGB}{193, 147, 6}
\definecolor{mylime}{RGB}{92, 136, 62}
\definecolor{myred}{RGB}{192, 0, 0}

\usepackage{hyperref}
\hypersetup{
    colorlinks=true,
    linkcolor=NavyBlue,
    filecolor=black,      
    urlcolor=black,
    citecolor=Gray,
    }

\usepackage{newfloat}
\usepackage{listings}
\DeclareCaptionStyle{ruled}{labelfont=normalfont,labelsep=colon,strut=off} % DO NOT CHANGE THIS
\floatstyle{ruled}
\newfloat{listing}{tb}{lst}{}
\floatname{listing}{Listing}
\title{ConVQG: Contrastive Visual Question Generation with Multimodal Guidance}

\author {
    Li Mi\equalcontrib,
    Syrielle Montariol\equalcontrib,
    Javiera Castillo-Navarro\equalcontrib,
    Xianjie Dai, \\
    Antoine Bosselut and Devis Tuia
}
\affiliations {
    EPFL, Switzerland \\
    \{li.mi, antoine.bosselut, devis.tuia\}@epfl.ch
}

\usepackage{bibentry}
\begin{document}

\maketitle

\begin{abstract}
Asking questions about visual environments is a crucial way for intelligent agents to understand rich multi-faceted scenes, raising the importance of Visual Question Generation (VQG) systems. Apart from being grounded to the image, existing VQG systems can use textual constraints, such as expected answers or knowledge triplets, to generate focused questions. These constraints allow VQG systems to specify the question content or leverage external commonsense knowledge that can not be obtained from the image content only. However, generating focused questions using textual constraints while enforcing a high relevance to the image content remains a challenge, as VQG systems often ignore one or both forms of grounding. In this work, we propose Contrastive Visual Question Generation (ConVQG), a method using a dual contrastive objective to discriminate questions generated using both modalities from those based on a single one. Experiments on both knowledge-aware and standard VQG benchmarks demonstrate that ConVQG outperforms the state-of-the-art methods and generates image-grounded, text-guided, and knowledge-rich questions. Our human evaluation results also show preference for ConVQG questions compared to non-contrastive baselines.
\end{abstract}

\section{Introduction}

Modern intelligent agents, like chatbots and dialog systems~\cite{ouyang2022training}, nowadays achieve (almost) human conversational skills, thanks to the development of large language models~\cite{brown2020language}. With the advances in vision-language research, we are now leaning towards visual dialog systems~\cite{das2017visual, openai2023gpt4}, which should be able to understand and interpret visual scenes and at the same time communicate with users. In this context, they should not only be able to provide answers but also be aware of what they do not know and request complementary information by asking questions about visual content.

Consequently, Visual Question Generation (VQG,~\cite{imvqg2019, zhang2017automatic}) becomes a rising area at the intersection of computer vision and natural language processing. VQG agents aim to generate meaningful and engaging questions for visual stimuli such as images. These images often depict multi-faceted scenes, with many salient elements that can be elaborated upon by asking focused questions.
\begin{figure}
	\centering
\includegraphics[width=0.46\textwidth]{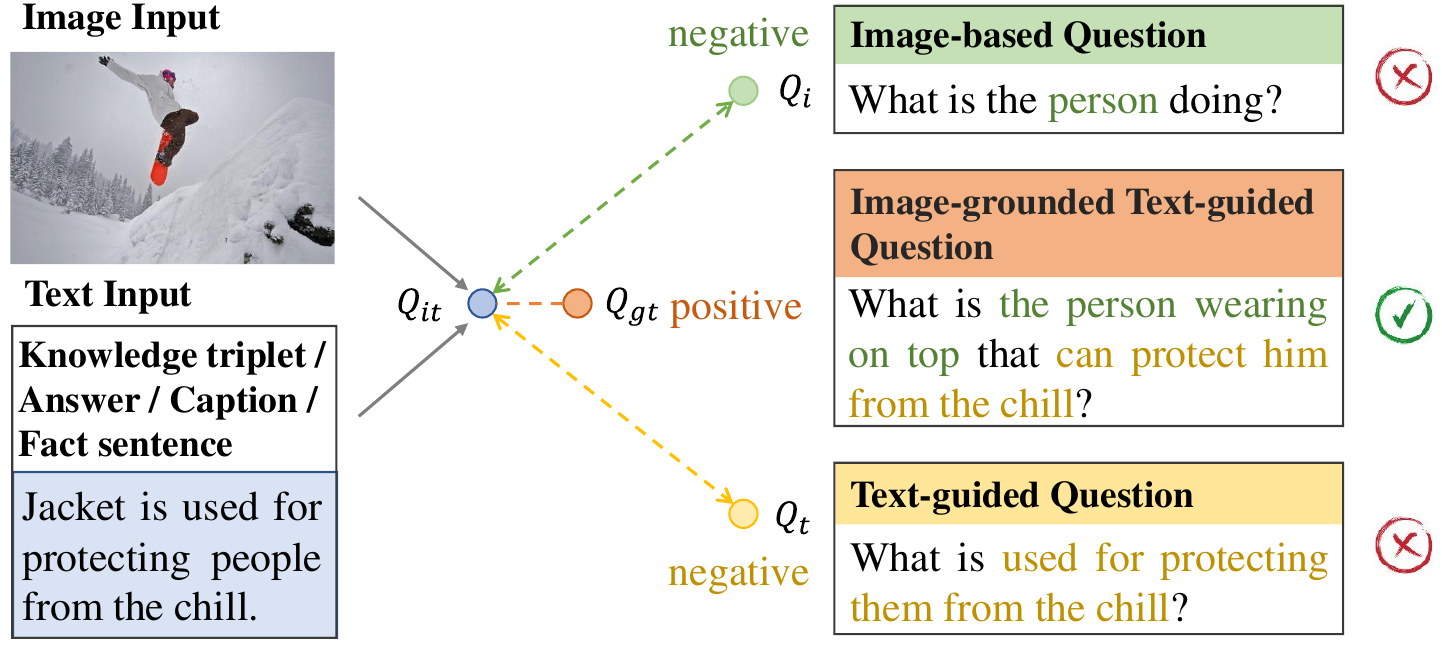}
	\caption{
 ConVQG at a glance. An image and a text input are processed through a multimodal module
, leading to the embedding $Q_{it}$. Pre-trained modules (detailed in Fig.~\ref{fig:framework}) produce image-only and text-only question embeddings ($Q_i$ and $Q_t$). A contrastive loss is then optimized to make $Q_{it}$ close to the real question embedding $Q_{gt}$ and far from the single modality ones. By design, ConVQG generates questions that are image-grounded (in \textcolor{mylime}{green}) and that meet the requirements of the text constraint (in \textcolor{myyellow}{yellow}).
}
	\label{fig:motivation}
\end{figure}

Early VQG systems tend to generate generic questions not exploiting the rich semantic content of the specific images. For example, the question ``\textit{What is the person doing?}" can be asked for any image containing a person. To make the question more focused, existing VQG systems exploit textual constraints, such as expected answers or knowledge triplets, as guidance. However, generating questions that are guided by a textual constraint while enforcing high relevance to the image content remains a challenge, since VQG systems often ignore one or both forms of grounding. 

To tackle these challenges, we propose \textbf{Con}trastive \textbf{V}isual \textbf{Q}uestion \textbf{G}eneration (\textbf{ConVQG}), a system that generates questions that (1) are based on details unique to a specific image, and (2) can be controlled using text to focus on specific objects, actions or concepts. 
To achieve that, the proposed method uses two modality-specific contrastive objectives to guide the generation of the question. The image contrastive objective drives the question away from a question generated using the image alone. The text contrastive objective drives the question away from one generated using only the textual constraint, enforcing more specific descriptions of the image while providing explicit control over the diversification of the generated questions. The textual constraint format is highly flexible; it can come from the answer to the question, a caption describing the image, or a knowledge triplet associated with an object or an action in the image. The latter, in particular, allows the model to enrich the generated question with image-grounded commonsense knowledge. These elements are found in existing public visual question-answering and question-generation datasets. Together, the two contrastive objectives allow the model to generate a diversified, rich and image-specific set of questions following textual constraints.

Through extensive experiments in standard and knowledge-aware VQG benchmarks, we show that ConVQG consistently outperforms state-of-the-art methods while providing flexibility regarding the type of textual constraints that can be used (answer, knowledge triplet or caption). Additionally, we perform a human evaluation using Amazon Mechanical Turk that shows the effectiveness of the contrastive learning objective to provide image-grounded and text-guided questions.

\section{Related Works}

\paragraph{Visual Question Generation.}
VQG is a particular case of question generation where the goal is to create one or several questions about a given image~\cite{zhang2017automatic}. 
Early VQG approaches focused on rules or template-based techniques~\cite{vijayakumar2016diverse, geman2015visual}. With the rise of neural networks, VQG was formulated as an image-to-sequence problem, designing an image encoder followed by a decoder to generate questions in natural language~\cite{ren2015exploring,grnn2016, li2018visual, patro2018multimodal}. However, these approaches often lead to poorly image-grounded and generic questions~\cite{kvqg2022, imvqg2019}. To avoid generic questions, \textit{text-guided VQG} has emerged, providing systems some guidance to obtain questions with specific properties. The constraint can be either the expected answer~\cite{radial2020, moag2021}, a question type~\cite{imvqg2019}, specific parts of the image~\cite{vedd2022guiding} or some external knowledge~\cite{kvqg2023}. In this work, we propose a VQG method to generate questions guided by text inputs (\textit{e.g.}, a knowledge triplet or the expected answer), which, together with our learning objective, ensures that the generated question is image-grounded and knowledge-aware.

\paragraph{Contrastive Learning (CL).}
The core idea of CL is learning by comparing. Given an anchor, CL defines a positive and a negative distribution, such that samples from the positive distribution (similar inputs) will be pulled together in the latent space while negative samples (dissimilar ones) will be pushed apart. CL has shown impressive performances on self-supervised and supervised learning in computer vision~\cite{chen2020simclr, he2020moco, khosla2020supcon}; natural language processing~\cite{oord2018representation, klein2021attention}, and audio processing~\cite{saeed2021contrastive} applications.
More recently, CL has shown remarkable results for multimodal embedding alignment in vision-language tasks~\cite{radford2021learning, jia2021scaling}. Indeed, contrastive objectives can be exploited to align representations of data pairs from different modalities (\textit{e.g.}, an image and its textual description). In this work, we leverage a contrastive objective to generate questions that consider visual and textual information together by learning a more distinguishable multimodal text-image joint representation from any single modality representation.

\paragraph{Vision-Language Pretraining (VLP).}

Benefiting from the success of language model pre-training~\cite{devlin2018bert, raffel2020exploring, brown2020language} and the recent development of model architectures in the communities~\cite{dosovitskiy2020image}, VLP boosts a large amount of vision-language tasks by providing a powerful vision-language joint representation~\cite{gan2022vision, chen2023vlp}. Those representations are usually pre-trained on large-scale datasets~\cite{schuhmann2021laion, lin2014microsoft} using simple objectives such as masked language modelling~\cite{devlin2018bert}, text-image matching~\cite{radford2021learning, jia2021scaling} or masked image modelling~\cite{chen2020uniter} and can be fine-tuned for various downstream vision-language tasks (\textit{e.g.}, text-image retrieval~\cite{kiros2014unifying}, image captioning~\cite{anderson2018bottom}, visual question answering~\cite{antol2015vqa}). 
In this paper, we build our baseline upon one of these models, BLIP~\cite{li2022blip}, for the powerful abilities provided by VLP. The proposed contrastive objectives serve as one way of tuning models for more readily accessing knowledge, while also distinguishing pure language commonsense from image-grounded ones.

\section{Contrastive Visual Question Generation}\label{sec:model}

\begin{figure*}
	\centering
	\includegraphics[width=0.97\textwidth]{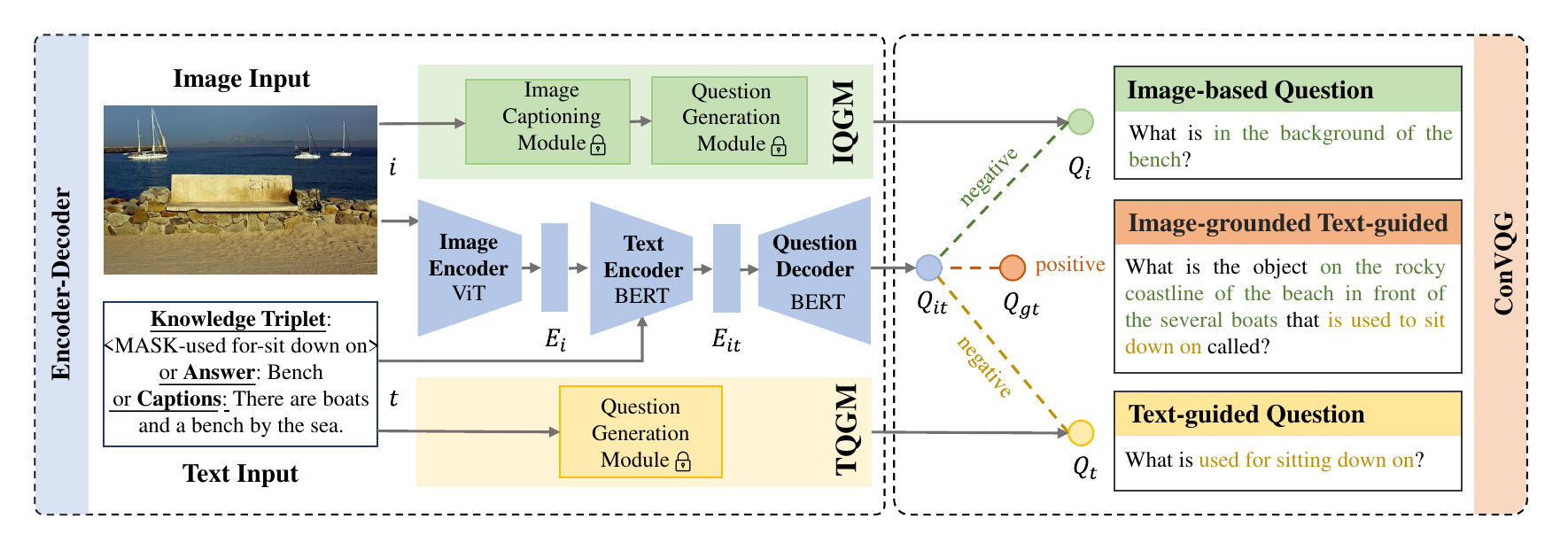}
	\caption{Pipeline of the ConVQG method. During training, an encoder-decoder VQG framework is powered by two additional branches for image-based question generation (IQGM) and text-based question generation (TQGM) (left part, the locker icon means the model is frozen). Then, contrastive losses discriminate image-text joint embeddings with the one from single modality only (right part). During inference, only the encoder-decoder framework is activated.}
	\label{fig:framework}
\end{figure*}

This section introduces our proposed visual question generation method, ConVQG, illustrated in Fig.~\ref{fig:framework}. In a nutshell, ConVQG is based on a multimodal encoder-decoder framework, trained in a contrastive way. The multimodal feature is contrasted against negative pairs obtained from single-modality generators to ensure that the generated question can not be obtained from a single modality alone.

\subsection{Problem Definition}
Given an image $i$, VQG aims at generating a reasonable and pertinent question $q$. On top of this, the question should meet a given requirement (\textit{e.g.}, reflecting constraints expressed by knowledge triplets or resulting in a given answer), which can be expressed as a text constraint $t$. The problem is solved by a multi-modal question generation model $p(q|i, t)$, which embeds image and text into a joint embedding and decodes a question based on image content and text constraints. 

\subsection{Architecture}\label{sec:vqg_framework}

ConVQG is built based on BLIP~\cite{li2022blip}, which is a large-scale vision-language pre-training pipeline consisting of an image encoder, a text encoder and a text decoder. Nevertheless, our proposed contrastive method can be used with any vision-language model.

\noindent \textbf{Image Encoder.} The image encoder is a vision transformer (ViT)~\cite{dosovitskiy2020image}. It receives an image $i$ as input, splits it into patches, and then feeds them into a transformer encoder~\cite{vaswani2017attention} to output a sequence of embeddings $E_i$: $E_{i} = \mathbf{ViT}(i)$.

\noindent \textbf{Text Encoder.} The text encoder of ConVQG is a variation of the BERT model~\cite{devlin2018bert} augmented with additional cross-attention layers at each transformer block to inject visual information into the text encoder. In this way, the text encoder takes as input both the image feature $E_i$ learned by the image encoder and some text $t$ constraining the question to be generated. Such text constraint can take various forms: a knowledge triplet (e.g, $<$\emph{MASK-used for-sit down on}$>$),\footnote{Here, the \textit{MASK} token replaces the answer to the question} a potential answer (\textit{e.g.}, $bench$), or any other information about the question or the image. They are formulated in natural language $t'$ (shown in supplementary materials). The output of the text encoder is regarded as a joint embedding of the image and text information $E_{it}$. The text encoder can be formulated as: $E_{it} = \mathbf{BERT}_{encoder}(t',E_{i})$.

\noindent \textbf{Question Decoder.} 
The ConVQG question decoder is analogous to the text decoder from BLIP. Essentially, it is a BERT model which replaces the bi-directional self-attention layers with causal self-attention ones. Thus, the inputs to the question decoder are the image-grounded text features learned by the text encoder while the output is the question embedding: $Q_{it} = \mathbf{BERT}_{decoder}(E_{it})$.

\subsection{Contrastive Learning for VQG}
A contrastive learning objective is proposed to generate the question based on both image and text information. The basic idea is that joint embeddings of images and text are supposed to be closer to the embeddings of the question annotations (\textit{i.e.}, the ground truth) while being different from those extracted from unimodal models considering the image (IQGM) or text (TQGM) in isolation.

\noindent \textbf{Image-based Question Generation Module (IQGM).} To generate questions based solely on visual information, we first use an image captioning model ($\mathbf{Cap}$) from BLIP to generate captions based on the image content. Then we use a question generation model~\cite{ushio2022generative} ($\mathbf{QG}$) to generate questions based on these captions. Finally, the generated questions are sent to a sentence-BERT model~\cite{sentencebert2019} to obtain the image-based question embeddings $Q_{i}$. The models are pre-trained. The IQGM can be denoted as Eq.~\eqref{IQGM}:
\begin{equation}
    Q_{i} = \mathbf{sBERT}(\mathbf{QG}(\mathbf{Cap}(i))).
    \label{IQGM}
\end{equation}

\noindent \textbf{Text-based Question Generation Module (TQGM).} 
The TQGM uses the same pre-trained question generation model~\cite{ushio2022generative} ($\mathbf{QG}$) as the IQGM, generating questions from the textual input processed as a sentence ($t'$). Then, the same sentence-BERT~\cite{sentencebert2019} model is used to embed the text-based question:
\begin{equation}
    Q_{t} = \mathbf{sBERT}(\mathbf{QG}(t')).
\end{equation}

\noindent \textbf{Contrastive Losses for VQG.}
To ensure VQG focuses both on image and text information, we propose a CL objective. With IQGM and TQGM, we obtain questions that are based only on visual information and text constraints respectively. Then we propose two contrastive losses, one on the image and one on the text. The image contrastive loss $CL_{img}$ enforces the L2-norm between the embedding generated by the IQGM, $Q_{it}$, and the embedding of the ground truth $Q_{gt}$, by the same sentence-BERT model, to be closer than the L2-norm between $Q_{it}$ and the image-only question embedding $Q_{i}$ by a margin $m$:
\begin{equation}
\medmath{
    CL_{img} = \max \left(\Vert Q_{it}-Q_{gt}\Vert_2 - \Vert Q_{it}-Q_{i}\Vert_2 +m, 0\right).}
    \label{eq:IQGM}
\end{equation}
The text-contrastive loss $CL_{txt}$ is analogous, using the embedding of the text-only model $Q_{t}$ as negative signal:
\begin{equation}
\medmath{
    CL_{txt} = \max \left(\Vert Q_{it}-Q_{gt} \Vert_2 - \Vert Q_{it}-Q_{t} \Vert_2 +m, 0\right).}
    \label{eq:TQGM}
\end{equation}
Then, the contrastive loss can be formulated as a weighted sum of $CL_{txt}$ and $CL_{img}$ with a parameter $\alpha$:
\begin{equation}
    CL = \alpha CL_{txt} + (1-\alpha) CL_{img}.
    \label{eq:CL}
\end{equation} 
Finally, the $CL$ loss is combined with a cross-entropy loss $CEL$ between predicted question embeddings and ground truth questions to ensure sufficient information from single modalities. The final loss of the ConVQG model can be represented as: 
\begin{equation}
Loss = (\beta CL + CEL)/2,
\label{eq:finalLoss}
\end{equation}
where $\beta$ is a parameter that can be fixed or tuned; it balances the contributions of the contrastive loss and the cross-entropy loss. In the Results section, we perform experiments to analyse the impact of these hyper-parameters.

\medskip

\noindent \textbf{Training and inference.} \label{sec:convqg_inference}
IQGM and TQGM are auxiliary, frozen modules. Therefore, the trainable components of ConVQG are only the image and text encoders of the multimodal branch, as well as the text decoder. At inference time, IQGM and TQGM are dropped, and only the multimodal encoder-decoder is used to obtain the question embedding $Q_{it}$. Then we use beam search, as in the sentence generator from BLIP, to decode the final question from $Q_{it}$.

\section{Experimental Setup}\label{sec:exp}

We compare ConVQG with several methods from the literature, considering different forms of \emph{text inputs}. In this section, we describe the datasets, metrics and the experimental settings that we used for training and evaluation.

\subsection{Datasets}

We evaluate our VQG method on three public datasets: a knowledge-aware benchmark (K-VQG) and two standard VQG benchmarks (VQA 2.0 and VQG COCO).

 \noindent \textbf{K-VQG}\footnote{\url{https://uehara-mech.github.io/kvqg}}~\cite{kvqg2023} is a knowledge-aware VQG dataset. It is a large-scale, humanly annotated dataset, where image-grounded questions are tied to structured knowledge (knowledge triplets). 
 Each sample consists of an image, a question, an answer, and a knowledge triplet. K-VQG contains $\medmath \sim$13K images and $\medmath \sim$16K (question, answer) pairs, related to $\medmath \sim$6K knowledge triplets.

\noindent \textbf{VQA 2.0}\footnote{\url{https://visualqa.org/download.html}}~\cite{goyal2017making} 
with more than 1M (image, question, answer) triplets, it is the largest and most commonly used dataset for VQG evaluation.
Images come from the COCO dataset~\cite{lin2014microsoft}, and three (question, answer) pairs were collected per image.
In our experiments, we consider two versions of this dataset: VQA 2.0 small~\cite{radial2020}, containing $\medmath \sim$80K images and $\medmath \sim$200K (question, answer) pairs; and VQA 2.0 large~\cite{imvqg2019}, which count $\medmath \sim$120K and $\medmath \sim$470K, respectively.

\noindent \textbf{VQG COCO}\footnote{\url{https://www.microsoft.com/en-us/download/details.aspx?id=53670}}~\cite{grnn2016} was created to generate natural and engaging questions for images. It contains 2500 training images, 1250 validation images, and 1250 testing images. Each image contains five natural questions and five ground truth captions. Different from the other two datasets, the answers are not always provided.

\subsection{Evaluation Metrics}
\label{sec:metrics}

\textbf{Numerical metrics.}
We use a variety of language generation metrics for evaluation: BLEU~\cite{papineni2002bleu}, METEOR~\cite{denkowski2014meteor} and CIDEr~\cite{vedantam2014cider}. They assess the conformity between questions generated by a model and ground truth questions. CIDEr, a TF-IDF-based metric, is the closest to human evaluation for image description compared to the other metrics~\cite{vedantam2014cider}. Additional information on how these metrics are computed can be found in the supplementary material. 
Similarly to most work in the literature~\cite{chen2015microsoft, moag2021}, we use the  \texttt{pycocoevalcap} package\footnote{\url{https://pypi.org/project/pycocoevalcap/}} for computing the metrics.

\noindent \textbf{Human evaluation.}
We use Amazon Mechanical Turk to assess the quality of \textit{model-generated} questions, asking workers to express their preferences about $500$ examples extracted from the K-VQG test set. 
Annotators must choose which of the two questions is better according to two criteria: (1) \textit{grounding to the knowledge triplet}, and (2) \textit{grounding to the image}. They also can indicate when none of the two questions is considered better if they consider that their similarity is too high to make a meaningful choice.
Additional details on the sample selection, the human evaluation process, and the instructions and examples given to the workers can be found in the supplementary material.

\subsection{Experimental Framework}

Following BLIP, the image encoder is a ViT-B/16, \textit{i.e}., a ViT architecture with 12 attention heads, 12 hidden layers, and images divided into $16\times16$ patches. The text encoder and the question decoder are $\text{BERT}_{\text{base}}$ models, \textit{i.e.}, transformer encoder with 12 attention heads and 12 hidden layers.
We initialize the encoder-decoder architecture with the corresponding pre-trained modules from BLIP~\cite{li2022blip}. Since all BLIP models are publicly available,\footnote{\url{https://github.com/salesforce/BLIP}} we choose the ``BLIP w/ ViT-B and CapFilt-L'' checkpoint for initialization. This model was pre-trained on 129M noisy image-text pairs using CapFilt-L, a captioning and filtering method.

Training was done on six NVIDIA A100-SXM4-40GB with a batch size of 24 each (VQA 2.0 dataset) and four NVIDIA V100-SXM2-32GB with a batch size of 16 each (K-VQG dataset, VQG-COCO dataset). The number of epochs varies depending on the dataset (10 for VQA 2.0, 5 for K-VQG, 5 for VQG-COCO). The starting learning rate is 2e-5 with a weight decay of 0.05.

\begin{table}[t]
  \centering
  \setlength{\tabcolsep}{3pt}
    \begin{tabular}{>{\centering \arraybackslash}m{0.19\linewidth}m{0.24\linewidth}>{\centering \arraybackslash}m{0.15\linewidth}>{\centering \arraybackslash}m{0.14\linewidth}>{\centering \arraybackslash}m{0.14\linewidth}}
    \toprule
          \emph{Text constraint} & \emph{Method} & {BLEU-4} & {METEOR} & {CIDEr} \\
    \midrule
    \multirow{2}{*}{Answer}
    &{IM-VQG \iffalse \cite{imvqg2019} \fi} & 12.37  & 16.65  & 0.39  \\
    &\textbf{ConVQG}$_{IT}$ & \textbf{14.30} & \textbf{18.67}  & \textbf{0.78} \\
    \midrule
    \multirow{1}{=}{\centering Knowledge Triplet}
    &{K-VQG \iffalse \cite{kvqg2023} \fi} & 18.84  & \textbf{22.79}  & 1.31  \\
    &\textbf{ConVQG}$_{IT}$ & \textbf{20.01} & 22.66  & \textbf{1.53} \\
    \bottomrule
    \end{tabular}%
    \caption{Results on the K-VQG dataset. The results of IM-VQG are reproduced based on the official code. The results of KVQG are taken from the respective paper.\protect\footnotemark
  }
  \label{tab:kvqg}%
\end{table}%

\section{Results} \label{sec:results}

In this section, we report the VQG results including quantitative, qualitative and human evaluation results. We compare ConVQG with several systems from the literature. For the sake of space, we report here only a subset of results from the literature. Additional results and descriptions of the competing methods can be found in the supplementary material.

\begin{table}[t]
  \centering
  \setlength{\tabcolsep}{2pt}
    \begin{tabular}{>{\centering \arraybackslash}m{0.16\linewidth}m{0.22\linewidth}>{\centering \arraybackslash}m{0.16\linewidth}>{\centering \arraybackslash}m{0.16\linewidth}>{\centering \arraybackslash}m{0.16\linewidth}}
    \toprule
          \emph{Test set} & \emph{Method} & {BLEU-4} & {METEOR} & {CIDEr} \\
          \midrule
          {\multirow{6}{*}{Small}} & {IVQA \iffalse \cite{liu2018inverse} \fi}  & {23.9} & {\textbf{35.7}} & {1.84 }  \\
             & {IM-VQG \iffalse \cite{imvqg2019} \fi}  & {24.8} & {26.3} & {1.94 }   \\
            & {iQAN \iffalse \cite{li2018visual} \fi} & {27.1} & {26.8}  & {2.09 }  \\
              & {Radial-GCN \iffalse \cite{radial2020} \fi}  & {27.9} & {27.1}  & {2.10 }  \\
               & {MOAG \iffalse \cite{moag2021} \fi} & {28.1} & {27.8} & {2.39 }  \\
               & {\textbf{ConVQG$_{IT}$}} & {\textbf{33.1}} & {30.0} & {\textbf{2.79}}  \\
          \midrule
          {\multirow{3}{*}{Large}} & {C3VQG \iffalse \cite{2021c3vqg} \fi} & {10.0 } & {13.6 }  & {0.47 }  \\
          & {IM-VQG \iffalse \cite{imvqg2019} \fi}  & {16.3 } & {20.6 } & {0.94 }  \\
               & {\textbf{ConVQG$_{IT}$}} & \textbf{22.4} & {\textbf{21.8}}  & {\textbf{1.78}}  \\
\bottomrule
    \end{tabular}%
    \caption{Results on the VQA 2.0 test sets. The results of the competing methods are taken from the respective papers.\protect \footnotemark} 
  \label{tab:vqa2}%
\end{table}%

\subsection{Results on VQG Benchmarks} \label{sec:res_quant}

\footnotetext{IM-VQG~\cite{imvqg2019}, K-VQG~\cite{kvqg2023}}
\footnotetext{IVQA~\cite{liu2018inverse}, IM-VQG~\cite{imvqg2019}, iQAN~\cite{li2018visual}, Radial-GCN~\cite{radial2020}, MOAG~\cite{moag2021}, C3VQG~\cite{2021c3vqg}}

We train ConVQG on three datasets, with different types of text inputs: knowledge triplets, answers and captions.

\noindent \textbf{Knowledge triplet.} Results are reported in Table~\ref{tab:kvqg} using the K-VQG dataset (row block \emph{Knowledge Triplet}), with masking the answers as in~\cite{kvqg2023}. ConVQG$_{IT}$ outperforms K-VQG \cite{kvqg2023} by 1.17\% on BLEU-4 and 0.22 points on CIDEr, and has a slightly lower METEOR score (0.13\% difference).

\noindent \textbf{Answer.} On the K-VQG dataset, answers can also be used as constraints. In Table~\ref{tab:kvqg} (row block \emph{Answer}), ConVQG$_{IT}$ shows an improvement of 1.93\% on BLEU-4, 2.02\% on METEOR and 0.39 points on CIDEr, with respect to the baseline method. On the VQA 2.0 dataset, samples consist of image, question, answer with no other additional sources of knowledge. Only the answer can be used as a text constraint.
Results on the VQA 2.0 dataset, large and small versions, are presented in Table~\ref{tab:vqa2}.
On the VQA 2.0 small, ConVQG$_{IT}$ leads to better performances for all the evaluation metrics.
The improvement on CIDEr (0.40 points) demonstrates that the generated questions become semantically similar to ground truth annotations. On VQA 2.0 large, ConVQG$_{IT}$ shows large improvements as well. Indeed, BLEU-4, METEOR, and CIDEr increased by 6.1\%, 1.2\%, and 0.84 points, respectively, with respect to SOTA approaches.

\noindent \textbf{Caption.} On the VQG-COCO dataset, there are no answers nor additional knowledge associated with questions, but captions are used as text inputs. We distinguish ConVQG$_{IT}^{*}$ from ConVQG$_{IT}$ because when captions are used as text constraints, the captioning step ($\mathbf{Cap}$) is skipped and questions generated by IQGM and TQGM are the same. Results show improvements among all metrics compared with the state-of-the-art methods. Compared with MC-BMN~\cite{patro2020deep}, BLEU-1, METEOR and CIDEr increase 9.5\%, 3.8\% and 0.06 points respectively (see Table~\ref{tab:vqgcoco}).

\begin{table}[!t]
\centering
\setlength{\tabcolsep}{4pt}
\begin{tabular}{m{0.22\linewidth}>{\centering \arraybackslash}m{0.18\linewidth}>{\centering \arraybackslash}m{0.18\linewidth}>{\centering \arraybackslash}m{0.18\linewidth}}
\toprule
\emph{Method} & {BLEU-1} & {METEOR}  & {CIDEr} \\
\midrule
MDN & 36.0 & 23.4  & 0.51 \\
MC-BMN & 40.7 & 22.6  & 0.50 \\
{\textbf{ConVQG$_{IT}^{*}$}} & \textbf{50.2} & \textbf{26.4} & \textbf{0.56} \\
\bottomrule
\end{tabular}%
\caption{Results on VQG-COCO, using captions as text constraint. We report BLEU-1 instead of BLEU-4 to be consistent with the comparison methods. The results for the competing methods are taken from the respective papers.\protect\footnotemark}
\label{tab:vqgcoco}%
\end{table}
\footnotetext{MDN~\cite{patro2018multimodal}, MC-BMN~\cite{patro2020deep}}

\subsection{Ablation Study}

\begin{figure*}[t]
	\centering
	\includegraphics[width=0.98\textwidth]{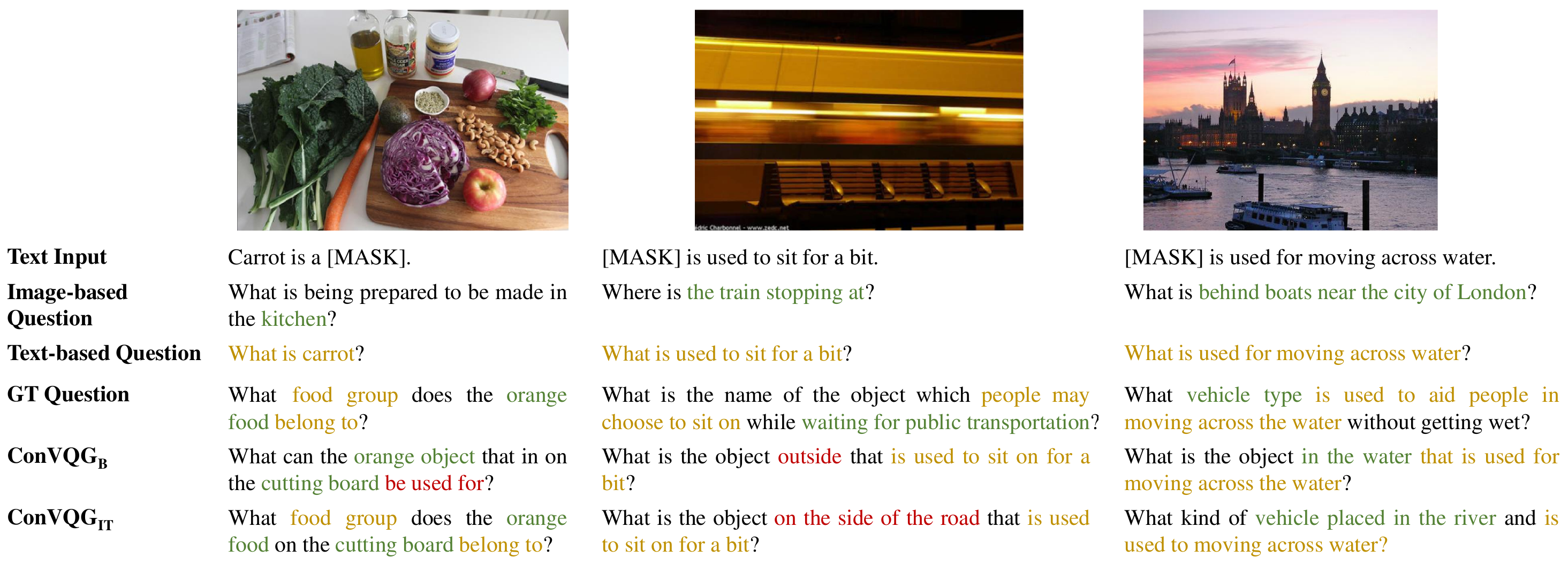}
	\caption{Examples from K-VQG dataset with knowledge triplets as inputs. In the text, \textcolor{mylime}{green color} denotes the sequence that is related to image content, while \textcolor{myyellow}{yellow color} denotes the information related to the text input. \textcolor{myred}{Red color} indicates wrong expressions, not related to the image nor the text input. Note: the raw input/output of the model is reported, without correcting grammar or syntax errors made by the generative model.}
	\label{fig:example_com}
\end{figure*}

In this section, we perform ablation studies to evaluate the contribution of each of the constrastive objectives. To this end, we distinguish four versions of our ConVQG model: 

\begin{enumerate}[partopsep=0pt, topsep=1pt, itemsep=0pt]
    \item \textbf{ConVQG}$_{B}$ is our baseline model, consisting of the multimodal encoder-decoder, without the contrastive modules, trained with cross-entropy loss.
    \item \textbf{ConVQG}$_{I}$ adds the IQGM module and the image contrastive loss in Eq.~\eqref{eq:IQGM} to the baseline model.
    \item \textbf{ConVQG}$_{T}$ adds the TQGM module and the text contrastive loss in Eq.~\eqref{eq:TQGM} to the baseline model.
    \item \textbf{ConVQG}$_{IT}$ is the full model as shown in Fig.~\ref{fig:framework}, that optimizes the final loss in Eq.~\eqref{eq:finalLoss}.
\end{enumerate}

Looking at the performance of ConVQG with some of its components deactivated (Table~\ref{tab:ablation}), we see that even the contrastive models using only the image (ConVQG$_I$) or text (ConVQG$_T$) contrastive module outperform the encoder-decoder baseline in all cases but one. For both cases, ConVQG$_{IT}$ works better than ConVQG$_{B}$, ConVQG$_{I}$ and ConVQG$_{T}$, especially for answers as inputs. ConVQG$_{IT}$ outperforms ConVQG$_{B}$ for 1.35\%, 0.89\% and 0.14 points on BLEU-4, METEOR and CIDEr respectively.

\begin{table}[t]
  \centering
  \setlength{\tabcolsep}{4pt}

    \begin{tabular}{>{\centering \arraybackslash}m{0.19\linewidth}m{0.24\linewidth}>{\centering \arraybackslash}m{0.15\linewidth}>{\centering \arraybackslash}m{0.14\linewidth}>{\centering \arraybackslash}m{0.14\linewidth}}
    \toprule
          \emph{Text constraint} & \emph{Method} & {BLEU-4} & {METEOR} & {CIDEr} \\
    \midrule
    \multirow{4}{*}{Answer}
    &{\textbf{ConVQG}$_{B}$} & 12.95 & 17.78 & 0.64 \\
    &{\textbf{ConVQG}$_{I}$} & 13.95 & 18.33 & 0.75 \\
    &{\textbf{ConVQG}$_{T}$} & 13.97 & 18.03  & 0.70 \\
    &\textbf{ConVQG}$_{IT}$ & \textbf{14.30} & \textbf{18.67}  & \textbf{0.78} \\
    \midrule
    \multirow{4}{=}{\centering Knowledge Triplet}
    &{\textbf{ConVQG}$_{B}$} & 18.33 & 21.47 & 1.31 \\
    &{\textbf{ConVQG}$_{I}$} & 19.00 & 21.91 & 1.38 \\
    &{\textbf{ConVQG}$_{T}$} & 19.11 & 20.65 & 1.39 \\
    &\textbf{ConVQG}$_{IT}$ & \textbf{20.01} & \textbf{22.66}  & \textbf{1.53} \\
    \bottomrule
    \end{tabular}%
      \caption{Ablation studies on K-VQG dataset.}
  \label{tab:ablation}%
\end{table}%

\label{sec:4.5.2}

\begin{table}[t]
  \centering
    \begin{tabular}{>{\centering \arraybackslash}m{0.1\linewidth}>{\centering \arraybackslash}m{0.1\linewidth}>{\centering \arraybackslash}m{0.15\linewidth}>{\centering \arraybackslash}m{0.15\linewidth}>{\centering \arraybackslash}m{0.15\linewidth}}
    \toprule
             \emph{Param.}  &  \emph{Value}     & {BLUE-4} & {METEOR} & {CIDEr} \\
          \midrule
          \multirow{2.5}[2]{*}{$\alpha$ }& {0.2} & {\textbf{20.01}} & {\textbf{22.66}} & {\textbf{1.53}}  \\
          & {0.5} & {19.90} & {22.60} & {1.52} \\
          & {0.8} & {19.79} & {22.56} & {1.52}  \\
          \midrule
          {\multirow{2.5}[2]{*}{$ \beta $}} & {10} & {19.80} & {22.55} & {1.52}  \\
              & {100} & {19.74} & {22.39} & {1.51}  \\
               & {Linear} & {\textbf{20.01}} & {\textbf{22.66}} & {\textbf{1.53}} \\
          \midrule
          {\multirow{2.5}[2]{*}{$m$}} & {0.2} & {19.89} & {\textbf{22.66}} & {\textbf{1.53}} \\
              & {0.5} & \textbf{20.01} & {\textbf{22.66}} & {\textbf{1.53}}  \\
              & {0.8} & {19.68} & {22.54} & {1.52} \\
          \bottomrule
    \end{tabular}%
    \caption{Parameter analysis on K-VQG dataset. $ \alpha$ from Eq.~\eqref{eq:CL}, $ \beta$ from Eq.~\eqref{eq:finalLoss} and $m$ from Eqs.~\eqref{eq:IQGM} and~\eqref{eq:TQGM}. Linear means $ \beta$ changed linearly during training.\protect \footnotemark}
  \label{tab:parameter}%
\end{table}%

\subsection{Parameter Analysis} 
In the proposed ConVQG method, there are three core parameters: 
$\alpha$ (Eq.~\eqref{eq:CL}), $\beta$  (Eq.~\eqref{eq:finalLoss}), both balancing the different parts of the loss and the margin $m$ (Eq.~\eqref{eq:IQGM} and \eqref{eq:TQGM}).
We vary their values and test their impact on ConVQG$_{IT}$ on the K-VQG dataset. Results are reported in Table~\ref{tab:parameter}.

All in all, these results show that ConVQG is robust to the model hyper-parameters since very small performance variations are observed. Linear $\beta$ outperforms fixed $\beta$ values, indicating that the contribution of the contrastive loss varies during training. $\alpha$ balances the relative contribution of image contrastive and text contrastive modules, which might vary depending on the dataset and how informative the text constraints are with respect to the image content. For $m$, metrics are relatively stable, especially for METEOR (max change 0.12\%) and CIDEr (max change 0.01 points).

\begin{figure}[t!]
    \centering
    \subfigure[One image - different text inputs]{
    \includegraphics[width=0.43\textwidth]{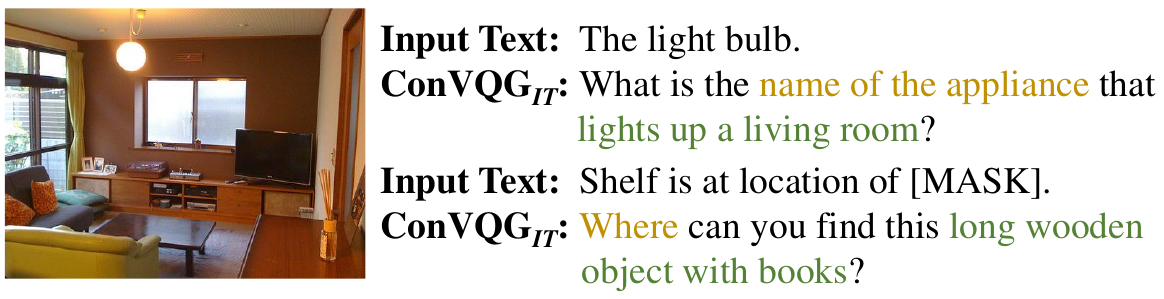}
	\label{fig:example_inputimage}
    }
    \subfigure[Different images - one text input]{
    \includegraphics[width=0.43\textwidth]{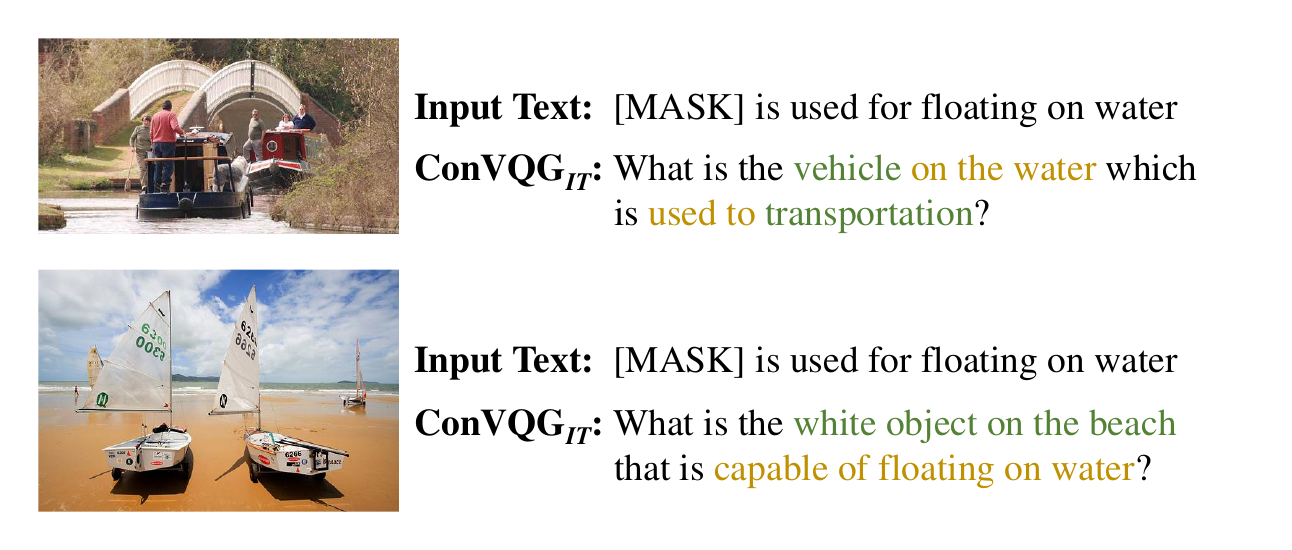}
	\label{fig:example_inputtext}
    }
    \caption{Question generation by ConVQG. Given the same image, it can generate different text-guided questions. Given the same text input, it can generate image-specific questions.}
\end{figure}

\subsection{Qualitative Results} 
Fig.~\ref{fig:example_com} shows generated questions on the K-VQG dataset. For each example, image and text inputs are displayed. Row  \emph{Image-based question} corresponds to the question generated by IQGM, while \emph{Text-based question} is the result obtained by TQGM. We compare questions generated by the proposed ConVQG$_{IT}$ with the outputs of the baseline without contrastive learning (ConVQG$_{B}$) and the annotations.

Comparing the questions generated by the ConVQG versions against the ground truth questions, we observe the following: first, ConVQG$_{IT}$ is able to constrain the question context according to the text inputs more precisely. For example, with the text constraint \emph{Carrot is a [Mask]}, the VQG model is supposed to generate a question about the category or a general description of \emph{Carrot}. The baseline method fails to understand the requirements behind the text input, while the proposed ConVQG$_{IT}$ generated a question that meets the constraint. Second, ConVQG$_{IT}$  provides more information based on both the visual scene (therefore referring to objects in the scene and their relationships) and the text context (formulated as a textual sentence). For instance, in the third example, ConVQG$_{IT}$ replaces \emph{in the water} (ConVQG$_{B}$) with a more precise generation of the image content (\emph{vehicle placed in the river}). We also provide failure cases, the models sometimes add inappropriate descriptions of images (the middle column) or fail to constrain the question with the text (ConVQG$_{B}$ in the first column).

\footnotetext{$\beta$ is increased by a factor of 10 at each epoch, starting from $\beta=10$.}

The ConVQG model can also be used in inference mode, where a single image and multiple knowledge triplets are given as inputs and vice versa. We show examples of both usages in Figs. \ref{fig:example_inputimage} and \ref{fig:example_inputtext}. \textbf{In the first case} (\emph{One image - different text inputs}, Fig.~\ref{fig:example_inputimage}), the generated questions capture the different constraints provided by the text input. For example, with answer \emph{The light bulb}, the model tries to describe it as \emph{lights up a living room} and \emph{has black and white stripes}. If the text input is changed to \emph{Shelf is at a location of [Mask]}, then the model generates a question about the place and adds more information such as \emph{long wooden object with books}. \textbf{In the second case}, if the model is given the same text and different images as inputs (\emph{Different images - one text input}, Fig.~\ref{fig:example_inputtext}), ConVQG$_{IT}$ generates image-grounded questions by finding unique image content. In the top example, ConVQG uses the words  \emph{vehicle} and \emph{transportation} in the question, showing general understanding provided by the visual cue of people traveling on the boats. In the bottom, the generated question contains the descriptions of the specific boats (\emph{white object}) and of the visual scene (\emph{on the beach}).

\subsection{Transfer Results} 
\begin{figure}[!t]
	\centering
\includegraphics[width=0.42\textwidth]{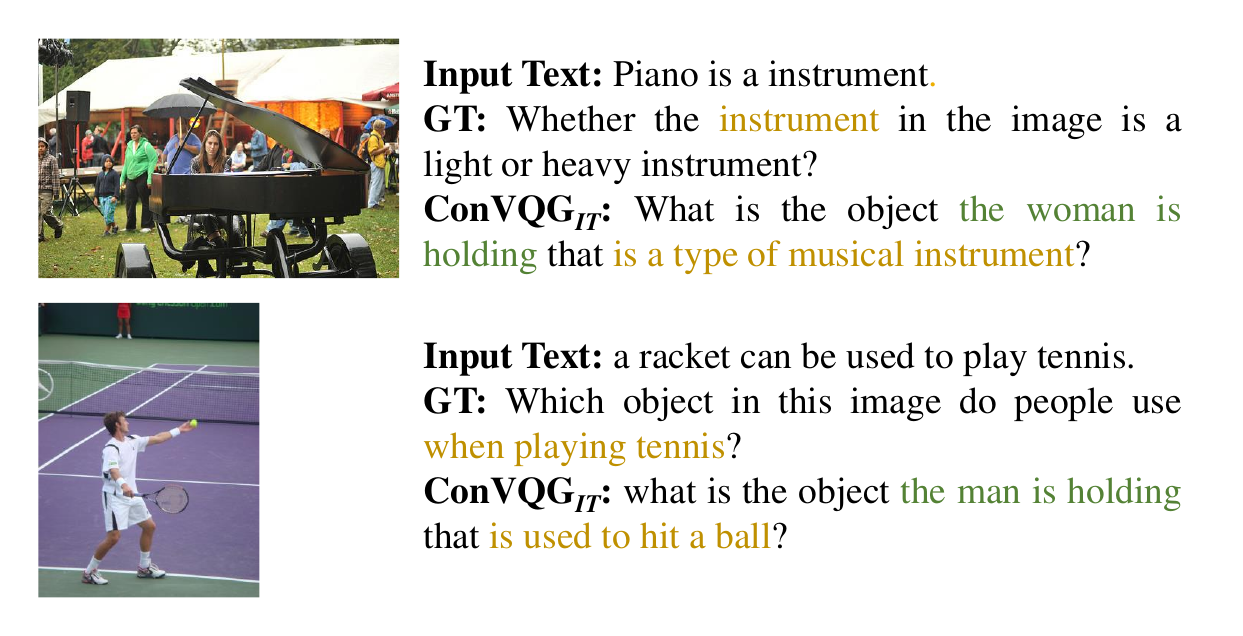}
	\caption{Transfer results on the FVQA dataset.}
	\label{fig:transfer}
\end{figure}

To demonstrate the generalization ability of ConVQG, we test it in a transfer setting: we train it on the K-VQG dataset and test it on the FVQA~\cite{wang2017fvqa} dataset without further training. FVQA was created for fact-based visual question answering.
 For each question-answer pair, a fact sentence is provided to clarify the possible commonsense to answer the question, which is used as a text constraint for our transfer settings.
Fig.~\ref{fig:transfer} illustrates this experiment. Compared with annotations, the question generated by ConVQG can be grounded to both image and text, which indicates the effectiveness of the contrastive objectives. Quantitative results can be found in supplementary materials.

\begin{figure}[!t]
    \centering    \includegraphics[width=0.37\textwidth]{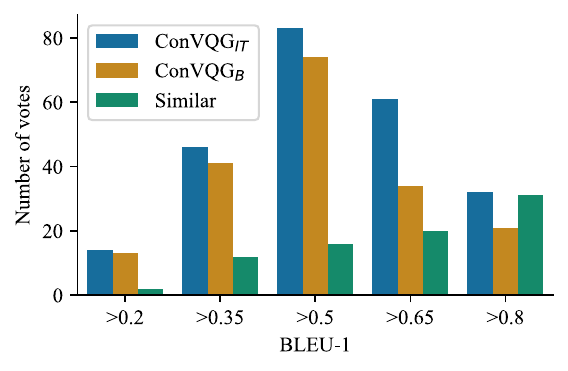}
    \caption{Histogram of human preference by similarity between the two questions, computed using BLEU-1 score.}
    \label{fig:mturl_res}

\end{figure}

\subsection{Human Evaluation Results} 
In this section, we report the results of the human evaluation performed on Amazon Mechanical Turk, on K-VQG test set. Among the 500 annotated question pairs, the question generated by {ConVQG}$_{IT}$  was preferred $236$ times; {ConVQG}$_{B}$ was preferred $183$ times; the option ``Similar'' was chosen $81$ times. We compute the similarity between the two questions using the BLEU-1 score. A histogram of the proportion of each of the three choices by degree of similarity between the questions can be found in Fig.~\ref{fig:mturl_res}. The proportion of the ``Similar'' option chosen by the annotators increases with the similarity between the questions, which is a good way to verify the ability of the workers to correctly tackle the task. Moreover, the contrastive model {ConVQG}$_{IT}$ is systematically chosen more often than the baseline model, demonstrating the human preference towards the proposed system.

\section{Conclusion}
Asking questions in natural language is a fundamental step toward effective visual dialog systems. In this work, we propose contrastive VQG with multimodal guidance from the image content and textual constraints. ConVQG leverages two modality-specific contrastive objectives to guide the content of the question by driving it away from questions generated from single modalities. Our multimodal system allows to control the diversity of questions, and the simultaneous grounding in both modalities. Extensive experiments in standard and knowledge-aware benchmarks show that ConVQG outperforms state-of-the-art methods and has good transfer capacities to unseen datasets. Human evaluation demonstrates that humans prefer ConVQG-generated questions to non-contrastive baselines. These results show that the contrastive objective of ConVQG is key to generating diverse, knowledge-rich, and image-specific questions.

\clearpage
\clearpage

\section*{Acknowledgements}
We thank the anonymous reviewers for their constructive and thoughtful comments. We also thank Siran Li and Chang Xu for providing codes of baselines; Zeming Chen, Tianqing Fang, Debjit Paul and Valérie Zermatten for providing helpful feedback on earlier versions of this work. We acknowledge the support from CSC and the EPFL Science Seed Fund. AB also gratefully acknowledges the support of the Swiss National Science Foundation (No. 215390), Innosuisse (PFFS-21-29), the EPFL Center for Imaging, Sony Group Corporation, and the Allen Institute for AI.

\bibliography{aaai24}

\begin{thebibliography}{58}
\providecommand{\natexlab}[1]{#1}

\bibitem[{Anderson et~al.(2018)Anderson, He, Buehler, Teney, Johnson, Gould, and Zhang}]{anderson2018bottom}
Anderson, P.; He, X.; Buehler, C.; Teney, D.; Johnson, M.; Gould, S.; and Zhang, L. 2018.
\newblock Bottom-up and top-down attention for image captioning and visual question answering.
\newblock In \emph{CVPR}, 6077--6086.

\bibitem[{Antol et~al.(2015)Antol, Agrawal, Lu, Mitchell, Batra, Zitnick, and Parikh}]{antol2015vqa}
Antol, S.; Agrawal, A.; Lu, J.; Mitchell, M.; Batra, D.; Zitnick, C.~L.; and Parikh, D. 2015.
\newblock VQA: Visual question answering.
\newblock In \emph{ICCV}, 2425--2433.

\bibitem[{Auer et~al.(2007)Auer, Bizer, Kobilarov, Lehmann, Cyganiak, and Ives}]{auer2007dbpedia}
Auer, S.; Bizer, C.; Kobilarov, G.; Lehmann, J.; Cyganiak, R.; and Ives, Z. 2007.
\newblock Dbpedia: A nucleus for a web of open data.
\newblock In \emph{ISWC}, 722--735.

\bibitem[{Brown et~al.(2020)Brown, Mann, Ryder, Subbiah, Kaplan, Dhariwal, Neelakantan, Shyam, Sastry, Askell et~al.}]{brown2020language}
Brown, T.; Mann, B.; Ryder, N.; Subbiah, M.; Kaplan, J.~D.; Dhariwal, P.; Neelakantan, A.; Shyam, P.; Sastry, G.; Askell, A.; et~al. 2020.
\newblock Language models are few-shot learners.
\newblock In \emph{NeurIPS}, 1877--1901.

\bibitem[{Chen et~al.(2023)Chen, Zhang, Han, Chen, Shi, Xu, and Xu}]{chen2023vlp}
Chen, F.-L.; Zhang, D.-Z.; Han, M.-L.; Chen, X.-Y.; Shi, J.; Xu, S.; and Xu, B. 2023.
\newblock {VLP}: A survey on vision-language pre-training.
\newblock \emph{Machine Intelligence Research}, 20(1): 38--56.

\bibitem[{Chen et~al.(2020{\natexlab{a}})Chen, Kornblith, Norouzi, and Hinton}]{chen2020simclr}
Chen, T.; Kornblith, S.; Norouzi, M.; and Hinton, G. 2020{\natexlab{a}}.
\newblock A simple framework for contrastive learning of visual representations.
\newblock In \emph{ICML}.

\bibitem[{Chen et~al.(2015)Chen, Fang, Lin, Vedantam, Gupta, Doll{\'a}r, and Zitnick}]{chen2015microsoft}
Chen, X.; Fang, H.; Lin, T.-Y.; Vedantam, R.; Gupta, S.; Doll{\'a}r, P.; and Zitnick, C.~L. 2015.
\newblock {Microsoft COCO captions: Data collection and evaluation server}.
\newblock \emph{arXiv preprint arXiv:1504.00325}.

\bibitem[{Chen et~al.(2020{\natexlab{b}})Chen, Li, Yu, El~Kholy, Ahmed, Gan, Cheng, and Liu}]{chen2020uniter}
Chen, Y.-C.; Li, L.; Yu, L.; El~Kholy, A.; Ahmed, F.; Gan, Z.; Cheng, Y.; and Liu, J. 2020{\natexlab{b}}.
\newblock {UNITER: Universal image-text representation learning}.
\newblock In \emph{ECCV}, 104--120.

\bibitem[{Das et~al.(2017)Das, Kottur, Gupta, Singh, Yadav, Moura, Parikh, and Batra}]{das2017visual}
Das, A.; Kottur, S.; Gupta, K.; Singh, A.; Yadav, D.; Moura, J.~M.; Parikh, D.; and Batra, D. 2017.
\newblock Visual dialog.
\newblock In \emph{CVPR}, 326--335.

\bibitem[{Deng et~al.(2009)Deng, Dong, Socher, Li, Li, and Fei-Fei}]{deng2009imagenet}
Deng, J.; Dong, W.; Socher, R.; Li, L.-J.; Li, K.; and Fei-Fei, L. 2009.
\newblock Imagenet: A large-scale hierarchical image database.
\newblock In \emph{CVPR}, 248--255.

\bibitem[{Denkowski and Lavie(2014)}]{denkowski2014meteor}
Denkowski, M.; and Lavie, A. 2014.
\newblock Meteor universal: Language specific translation evaluation for any target language.
\newblock In \emph{WMT}, 376--380.

\bibitem[{Devlin et~al.(2018)Devlin, Chang, Lee, and Toutanova}]{devlin2018bert}
Devlin, J.; Chang, M.-W.; Lee, K.; and Toutanova, K. 2018.
\newblock BERT: Pre-training of deep bidirectional transformers for language understanding.
\newblock \emph{arXiv preprint arXiv:1810.04805}.

\bibitem[{Dosovitskiy et~al.(2021)Dosovitskiy, Beyer, Kolesnikov, Weissenborn, Zhai, Unterthiner, Dehghani, Minderer, Heigold, Gelly et~al.}]{dosovitskiy2020image}
Dosovitskiy, A.; Beyer, L.; Kolesnikov, A.; Weissenborn, D.; Zhai, X.; Unterthiner, T.; Dehghani, M.; Minderer, M.; Heigold, G.; Gelly, S.; et~al. 2021.
\newblock An image is worth 16x16 words: Transformers for image recognition at scale.
\newblock In \emph{ICLR}.

\bibitem[{Gan et~al.(2022)Gan, Li, Li, Wang, Liu, Gao et~al.}]{gan2022vision}
Gan, Z.; Li, L.; Li, C.; Wang, L.; Liu, Z.; Gao, J.; et~al. 2022.
\newblock Vision-language pre-training: Basics, recent advances, and future trends.
\newblock \emph{Foundations and Trends{\textregistered} in Computer Graphics and Vision}, 14(3--4): 163--352.

\bibitem[{Geman et~al.(2015)Geman, Geman, Hallonquist, and Younes}]{geman2015visual}
Geman, D.; Geman, S.; Hallonquist, N.; and Younes, L. 2015.
\newblock {Visual Turing test for computer vision systems}.
\newblock \emph{Proceedings of the National Academy of Sciences}, 112(12): 3618--3623.

\bibitem[{Goyal et~al.(2017)Goyal, Khot, Summers-Stay, Batra, and Parikh}]{goyal2017making}
Goyal, Y.; Khot, T.; Summers-Stay, D.; Batra, D.; and Parikh, D. 2017.
\newblock Making the v in VQA matter: Elevating the role of image understanding in visual question answering.
\newblock In \emph{CVPR}, 6904--6913.

\bibitem[{He et~al.(2020)He, Fan, Wu, Xie, and Girshick}]{he2020moco}
He, K.; Fan, H.; Wu, Y.; Xie, S.; and Girshick, R. 2020.
\newblock Momentum contrast for unsupervised visual representation learning.
\newblock In \emph{CVPR}, 9729--9738.

\bibitem[{Jain, Zhang, and Schwing(2017)}]{jain2017creativity}
Jain, U.; Zhang, Z.; and Schwing, A.~G. 2017.
\newblock Creativity: Generating diverse questions using variational autoencoders.
\newblock In \emph{CVPR}, 6485--6494.

\bibitem[{Jia et~al.(2021)Jia, Yang, Xia, Chen, Parekh, Pham, Le, Sung, Li, and Duerig}]{jia2021scaling}
Jia, C.; Yang, Y.; Xia, Y.; Chen, Y.-T.; Parekh, Z.; Pham, H.; Le, Q.; Sung, Y.-H.; Li, Z.; and Duerig, T. 2021.
\newblock Scaling up visual and vision-language representation learning with noisy text supervision.
\newblock In \emph{ICML}.

\bibitem[{Karpathy and Fei-Fei(2015)}]{karpathy2015deep}
Karpathy, A.; and Fei-Fei, L. 2015.
\newblock Deep visual-semantic alignments for generating image descriptions.
\newblock In \emph{CVPR}, 3128--3137.

\bibitem[{Khosla et~al.(2020)Khosla, Teterwak, Wang, Sarna, Tian, Isola, Maschinot, Liu, and Krishnan}]{khosla2020supcon}
Khosla, P.; Teterwak, P.; Wang, C.; Sarna, A.; Tian, Y.; Isola, P.; Maschinot, A.; Liu, C.; and Krishnan, D. 2020.
\newblock Supervised contrastive learning.
\newblock In \emph{NeurIPS}, 18661--18673.

\bibitem[{Kiros, Salakhutdinov, and Zemel(2014)}]{kiros2014unifying}
Kiros, R.; Salakhutdinov, R.; and Zemel, R.~S. 2014.
\newblock Unifying visual-semantic embeddings with multimodal neural language models.
\newblock \emph{arXiv preprint arXiv:1411.2539}.

\bibitem[{Klein and Nabi(2021)}]{klein2021attention}
Klein, T.; and Nabi, M. 2021.
\newblock Attention-based contrastive learning for winograd schemas.
\newblock In \emph{EMNLP-Findings}, 2428--2434.

\bibitem[{Krishna, Bernstein, and Fei-Fei(2019)}]{imvqg2019}
Krishna, R.; Bernstein, M.; and Fei-Fei, L. 2019.
\newblock Information maximizing visual question generation.
\newblock In \emph{CVPR}, 2008--2018.

\bibitem[{Li et~al.(2022)Li, Li, Xiong, and Hoi}]{li2022blip}
Li, J.; Li, D.; Xiong, C.; and Hoi, S. 2022.
\newblock {BLIP: Bootstrapping language-image pre-training for unified vision-language understanding and generation}.
\newblock In \emph{ICML}.

\bibitem[{Li et~al.(2018)Li, Duan, Zhou, Chu, Ouyang, Wang, and Zhou}]{li2018visual}
Li, Y.; Duan, N.; Zhou, B.; Chu, X.; Ouyang, W.; Wang, X.; and Zhou, M. 2018.
\newblock Visual question generation as dual task of visual question answering.
\newblock In \emph{CVPR}, 6116--6124.

\bibitem[{Lin(2004)}]{lin2004rouge}
Lin, C.-Y. 2004.
\newblock Rouge: A package for automatic evaluation of summaries.
\newblock In \emph{Text summarization branches out}, 74--81.

\bibitem[{Lin et~al.(2014)Lin, Maire, Belongie, Hays, Perona, Ramanan, Doll{\'a}r, and Zitnick}]{lin2014microsoft}
Lin, T.-Y.; Maire, M.; Belongie, S.; Hays, J.; Perona, P.; Ramanan, D.; Doll{\'a}r, P.; and Zitnick, C.~L. 2014.
\newblock {Microsoft COCO: Common objects in context}.
\newblock In \emph{ECCV}, 740--755.

\bibitem[{Liu et~al.(2018)Liu, Xiang, Hospedales, Yang, and Sun}]{liu2018inverse}
Liu, F.; Xiang, T.; Hospedales, T.~M.; Yang, W.; and Sun, C. 2018.
\newblock {Inverse visual question answering: A new benchmark and VQA diagnosis tool}.
\newblock \emph{IEEE Transactions on Pattern Analysis and Machine Intelligence}, 42(2): 460--474.

\bibitem[{Mostafazadeh et~al.(2016)Mostafazadeh, Misra, Devlin, Mitchell, He, and Vanderwende}]{grnn2016}
Mostafazadeh, N.; Misra, I.; Devlin, J.; Mitchell, M.; He, X.; and Vanderwende, L. 2016.
\newblock Generating natural questions about an image.
\newblock In \emph{ACL}, 1802--1813.

\bibitem[{Oord, Li, and Vinyals(2018)}]{oord2018representation}
Oord, A. v.~d.; Li, Y.; and Vinyals, O. 2018.
\newblock Representation learning with contrastive predictive coding.
\newblock \emph{arXiv preprint arXiv:1807.03748}.

\bibitem[{OpenAI(2023)}]{openai2023gpt4}
OpenAI. 2023.
\newblock {GPT-4} Technical report.
\newblock Technical report.

\bibitem[{Ouyang et~al.(2022)Ouyang, Wu, Jiang, Almeida, Wainwright, Mishkin, Zhang, Agarwal, Slama, Ray et~al.}]{ouyang2022training}
Ouyang, L.; Wu, J.; Jiang, X.; Almeida, D.; Wainwright, C.; Mishkin, P.; Zhang, C.; Agarwal, S.; Slama, K.; Ray, A.; et~al. 2022.
\newblock Training language models to follow instructions with human feedback.
\newblock In \emph{NeurIPS}, 27730--27744.

\bibitem[{Papineni et~al.(2002)Papineni, Roukos, Ward, and Zhu}]{papineni2002bleu}
Papineni, K.; Roukos, S.; Ward, T.; and Zhu, W.-J. 2002.
\newblock BLEU: A method for automatic evaluation of machine translation.
\newblock In \emph{ACL}, 311--318.

\bibitem[{Patro et~al.(2020)Patro, Kurmi, Kumar, and Namboodiri}]{patro2020deep}
Patro, B.; Kurmi, V.; Kumar, S.; and Namboodiri, V. 2020.
\newblock Deep bayesian network for visual question generation.
\newblock In \emph{WACV}, 1566--1576.

\bibitem[{Patro et~al.(2018)Patro, Kumar, Kurmi, and Namboodiri}]{patro2018multimodal}
Patro, B.~N.; Kumar, S.; Kurmi, V.~K.; and Namboodiri, V.~P. 2018.
\newblock Multimodal differential network for visual question generation.
\newblock In \emph{EMNLP}, 4002--4012.

\bibitem[{Radford et~al.(2021)Radford, Kim, Hallacy, Ramesh, Goh, Agarwal, Sastry, Askell, Mishkin, Clark et~al.}]{radford2021learning}
Radford, A.; Kim, J.~W.; Hallacy, C.; Ramesh, A.; Goh, G.; Agarwal, S.; Sastry, G.; Askell, A.; Mishkin, P.; Clark, J.; et~al. 2021.
\newblock Learning transferable visual models from natural language supervision.
\newblock In \emph{ICML}.

\bibitem[{Raffel et~al.(2020)Raffel, Shazeer, Roberts, Lee, Narang, Matena, Zhou, Li, and Liu}]{raffel2020exploring}
Raffel, C.; Shazeer, N.; Roberts, A.; Lee, K.; Narang, S.; Matena, M.; Zhou, Y.; Li, W.; and Liu, P.~J. 2020.
\newblock Exploring the limits of transfer learning with a unified text-to-text transformer.
\newblock \emph{The Journal of Machine Learning Research}, 21(1): 5485--5551.

\bibitem[{Reimers and Gurevych(2019)}]{sentencebert2019}
Reimers, N.; and Gurevych, I. 2019.
\newblock {Sentence-BERT: Sentence embeddings using siamese BERT-networks}.
\newblock In \emph{EMNLP-IJCNLP}, 3982--3992.

\bibitem[{Ren, Kiros, and Zemel(2015)}]{ren2015exploring}
Ren, M.; Kiros, R.; and Zemel, R. 2015.
\newblock Exploring models and data for image question answering.
\newblock In \emph{NeurIPS}.

\bibitem[{Saeed, Grangier, and Zeghidour(2021)}]{saeed2021contrastive}
Saeed, A.; Grangier, D.; and Zeghidour, N. 2021.
\newblock Contrastive learning of general-purpose audio representations.
\newblock In \emph{ICASSP}, 3875--3879.

\bibitem[{Schuhmann et~al.(2021)Schuhmann, Vencu, Beaumont, Kaczmarczyk, Mullis, Katta, Coombes, Jitsev, and Komatsuzaki}]{schuhmann2021laion}
Schuhmann, C.; Vencu, R.; Beaumont, R.; Kaczmarczyk, R.; Mullis, C.; Katta, A.; Coombes, T.; Jitsev, J.; and Komatsuzaki, A. 2021.
\newblock LAION-400M: Open dataset of clip-filtered 400 million image-text pairs.
\newblock \emph{arXiv preprint arXiv:2111.02114}.

\bibitem[{Speer, Chin, and Havasi(2017)}]{speer2016conceptnet}
Speer, R.; Chin, J.; and Havasi, C. 2017.
\newblock {ConceptNet 5.5: An open multilingual graph of general knowledge}.
\newblock In \emph{AAAI}.

\bibitem[{Tandon et~al.(2014)Tandon, De~Melo, Suchanek, and Weikum}]{tandon2014webchild}
Tandon, N.; De~Melo, G.; Suchanek, F.; and Weikum, G. 2014.
\newblock Webchild: Harvesting and organizing commonsense knowledge from the web.
\newblock In \emph{WSDM}, 523--532.

\bibitem[{Uehara and Harada(2023)}]{kvqg2023}
Uehara, K.; and Harada, T. 2023.
\newblock K-{VQG}: Knowledge-aware visual question generation for common-sense acquisition.
\newblock In \emph{WACV}, 4401--4409.

\bibitem[{Uppal et~al.(2021)Uppal, Madan, Bhagat, Yu, and Shah}]{2021c3vqg}
Uppal, S.; Madan, A.; Bhagat, S.; Yu, Y.; and Shah, R.~R. 2021.
\newblock {C3VQG: Category consistent cyclic visual question generation}.
\newblock In \emph{ACM MM Asia}.

\bibitem[{Ushio, Alva-Manchego, and Camacho-Collados(2022)}]{ushio2022generative}
Ushio, A.; Alva-Manchego, F.; and Camacho-Collados, J. 2022.
\newblock Generative language models for paragraph-level question generation.
\newblock In \emph{EMNLP}, 670--688.

\bibitem[{Vaswani et~al.(2017)Vaswani, Shazeer, Parmar, Uszkoreit, Jones, Gomez, Kaiser, and Polosukhin}]{vaswani2017attention}
Vaswani, A.; Shazeer, N.; Parmar, N.; Uszkoreit, J.; Jones, L.; Gomez, A.~N.; Kaiser, {\L}.; and Polosukhin, I. 2017.
\newblock Attention is all you need.
\newblock In \emph{NeurIPS}.

\bibitem[{Vedantam, Lawrence~Zitnick, and Parikh(2015)}]{vedantam2014cider}
Vedantam, R.; Lawrence~Zitnick, C.; and Parikh, D. 2015.
\newblock {CIDEr: Consensus-based image description evaluation}.
\newblock In \emph{CVPR}, 4566--4575.

\bibitem[{Vedd et~al.(2022)Vedd, Wang, Rei, Miao, and Specia}]{vedd2022guiding}
Vedd, N.; Wang, Z.; Rei, M.; Miao, Y.; and Specia, L. 2022.
\newblock Guiding visual question generation.
\newblock In \emph{ACL}, 1640--1654.

\bibitem[{Vijayakumar et~al.(2016)Vijayakumar, Cogswell, Selvaraju, Sun, Lee, Crandall, and Batra}]{vijayakumar2016diverse}
Vijayakumar, A.~K.; Cogswell, M.; Selvaraju, R.~R.; Sun, Q.; Lee, S.; Crandall, D.; and Batra, D. 2016.
\newblock Diverse beam search: Decoding diverse solutions from neural sequence models.
\newblock \emph{arXiv preprint arXiv:1610.02424}.

\bibitem[{Wang et~al.(2017)Wang, Wu, Shen, Dick, and Van Den~Hengel}]{wang2017fvqa}
Wang, P.; Wu, Q.; Shen, C.; Dick, A.; and Van Den~Hengel, A. 2017.
\newblock FVQA: Fact-based visual question answering.
\newblock \emph{IEEE transactions on pattern analysis and machine intelligence}, 40(10): 2413--2427.

\bibitem[{Xie et~al.(2021)Xie, Cai, Huang, and Wang}]{moag2021}
Xie, J.; Cai, Y.; Huang, Q.; and Wang, T. 2021.
\newblock Multiple objects-aware visual question generation.
\newblock In \emph{ACM MM}, 4546--4554.

\bibitem[{Xie et~al.(2022)Xie, Fang, Cai, Huang, and Li}]{kvqg2022}
Xie, J.; Fang, W.; Cai, Y.; Huang, Q.; and Li, Q. 2022.
\newblock Knowledge-based visual question generation.
\newblock \emph{IEEE Transactions on Circuits and Systems for Video Technology}, 32(11): 7547--7558.

\bibitem[{Xu et~al.(2015)Xu, Ba, Kiros, Cho, Courville, Salakhudinov, Zemel, and Bengio}]{xu2015show}
Xu, K.; Ba, J.; Kiros, R.; Cho, K.; Courville, A.; Salakhudinov, R.; Zemel, R.; and Bengio, Y. 2015.
\newblock Show, attend and tell: Neural image caption generation with visual attention.
\newblock In \emph{ICML}.

\bibitem[{Xu et~al.(2018)Xu, Song, Lu, He, Yang, and Shen}]{xu2018dual}
Xu, X.; Song, J.; Lu, H.; He, L.; Yang, Y.; and Shen, F. 2018.
\newblock Dual learning for visual question generation.
\newblock In \emph{ICME}.

\bibitem[{Xu et~al.(2020)Xu, Wang, Yang, Hanjalic, and Shen}]{radial2020}
Xu, X.; Wang, T.; Yang, Y.; Hanjalic, A.; and Shen, H.~T. 2020.
\newblock Radial graph convolutional network for visual question generation.
\newblock \emph{IEEE Transactions on Neural Networks and Learning Systems}, 32(4): 1654--1667.

\bibitem[{Zhang et~al.(2017)Zhang, Qu, You, Yang, and Zhang}]{zhang2017automatic}
Zhang, S.; Qu, L.; You, S.; Yang, Z.; and Zhang, J. 2017.
\newblock Automatic generation of grounded visual questions.
\newblock In \emph{IJCAI}, 4235--4243.

\end{thebibliography}

\newpage

\section{Supplementary Materials}
\section{Datasets and Preprocessing}
\subsection{Datasets details}

In this section, we introduce more details about the datasets used for the evaluation of ConVQG.

\paragraph{K-VQG~\cite{kvqg2023}} is a knowledge-aware VQG dataset. It is the first large, human-annotated dataset in which image-grounded questions are tied to structured knowledge. To build the dataset, knowledge triplets were collected from two sources: ConceptNet and \textsc{Atomic}$_{20}^{20}$.

ConceptNet contains $\sim$34M triples and 37 types of relations, which are not all well-suited for image description; therefore, only 15 types of relations were selected as suitable targets for image-grounded questions. \textsc{Atomic}$_{20}^{20}$ contains $\sim$1M knowledge triplets, among which only physical-entity relations were retained for VQG. Both knowledge bases were then post-processed, giving a total of $\sim$150K knowledge triplets as candidate knowledge for VQG.

The question collection for K-VQG dataset was performed using Amazon Mechanical Turk (MTurk). The workers were given an image, the bounding box of a target object in the image, the name of the target object, and a list of candidate knowledge triplets. The workers were then asked to write knowledge-aware questions for the image by first selecting an appropriate knowledge triplet and an entity of the knowledge triplet that would be the answer to the question. Finally, an independent phase of question validation was performed on MTurk to ensure the quality of the collected questions.

Each sample in the dataset consists of an image, a question, an answer, a knowledge triplet, and a bounding box of the question target. As a result, K-VQG contains 13648 images and 16098 pairs, related to 6084 knowledge triplets.

In our experiments, we use the same dataset splits as in the original paper. 

\paragraph{VQA 2.0~\cite{goyal2017making}} is the most commonly used dataset for VQG evaluation~\cite{imvqg2019, moag2021}.
In particular, VQA 2.0 builds on top of the VQA dataset, which contains 204K images from COCO, 614K free-form natural language questions (3 per image), and over 6M free-form concise answers (10 per question).

Despite the significant progress the VQA dataset enabled in the field, it has been shown that language carries strong priors that can result in good superficial performance~\cite{goyal2017making}, even when models do not attend to the visual content. The questions and answers in VQA 2.0 have been carefully curated to alleviate these language biases. The idea is that for every \textit{(image, question, answer)} triplet $(I,Q,A)$ in the VQA dataset, one can find an image $I’$ (similar to $I$) that results in an answer $A’$ (different from $A$) to the same question $Q$.

MTurk is used to collect human-annotated data in two steps: (i) finding the complementary images $I’$, and (ii) collecting answers to the complementary $(I’, Q)$ image question pairs.
Thus, the VQA 2.0 contains more than 1M \textit{(image, question, answer)} triples, being the largest dataset for VQG evaluation to date.

Works in the literature have used the VQA 2.0 dataset with different train, validation, and test splits. For this reason, we consider two versions of this dataset to report our results: VQA 2.0 small~\cite{radial2020} and VQA 2.0 large~\cite{imvqg2019}. 
Additional information about these two versions can be found in Section Data preprocessing.

\paragraph{VQG-COCO~\cite{grnn2016}} was collected by selecting 5,000 images that were also annotated by CQA
dataset (Ren et al., 2015) and by VQA (Antol et
al., 2015), from the MS-COCO dataset~\cite{lin2014microsoft}. The main objective of constructing this dataset is to generate more natural and creative questions. The VQG-COCO dataset contains a total of 2500 training images, 1250 validation images, and 1250 testing images. For each image in the dataset, there are five natural questions and five ground truth captions.

\paragraph{FVQA~\cite{wang2017fvqa}} was created for fact-based visual question answering; this means that questions in the dataset need the support of some commonsense knowledge to be answered. 

To build the dataset, the authors first collected images from the COCO~\cite{lin2014microsoft} validation set and ImageNet~\cite{deng2009imagenet} test set. Three types of visual concepts were extracted from these images: objects, scene and action. Then, supporting facts were selected from knowledge bases, including ConceptNet~\cite{speer2016conceptnet}, DBpedia~\cite{auer2007dbpedia}, and WebChild~\cite{tandon2014webchild}. Knowledge triplets used from DBpedia concern categories and super-categories; ConceptNet relationships encode commonsense knowledge, while knowledge from WebChild encodes comparative relations.
During the question collection phase, human annotators were asked to provide visual questions that required a supporting fact to be answered.
FVQA contains 2190 images and 5826 \textit{(question, answer)} pairs. However, questions in this dataset have been criticized for being poorly grounded to the image \cite{goyal2017making}. For this reason, we only use FVQA for the transfer setting of ConVQG. Even though the results need to be taken with a pinch of salt.

More details about the datasets' splits used in this work can be found in Table~\ref{tab:datasets}.

\begin{table}[!htpb]
    \centering
    \setlength{\tabcolsep}{1.4pt}
    \begin{tabular}{>{\centering \arraybackslash}m{0.10\linewidth}>{\centering \arraybackslash}m{0.10\linewidth}>{\centering \arraybackslash}m{0.15\linewidth}>{\centering \arraybackslash}m{0.15\linewidth}>{\centering \arraybackslash}m{0.14\linewidth}>{\centering \arraybackslash}m{0.14\linewidth}>{\centering \arraybackslash}m{0.14\linewidth}}
    \toprule
     \multicolumn{2}{c}{Dataset} & VQA 2.0 small & VQA 2.0 large
     & K-VQG & VQG-COCO & FVQA \\\midrule
   \multirow{2}{*}{Train} & \cellcolor{Gray} \textit{QA} & \cellcolor{Gray}221\,708 & \cellcolor{Gray}294\,296
    & \cellcolor{Gray}12\,888 & \cellcolor{Gray} 12\,500 & \cellcolor{Gray} - \\
   & \textit{Img} & 76\,238 & 80\,630
   & 10\,915 & 2\,500 & - \\
   \multirow{2}{*}{Test} & \cellcolor{Gray} \textit{QA} & \cellcolor{Gray}12\,940 & \cellcolor{Gray}176\,868
   &\cellcolor{Gray} 3\,207 & \cellcolor{Gray} 6\,250 & \cellcolor{Gray} - \\
   & \textit{Img} & 4\,593 & 40\,305
   & 2\,730 & 1\,250 & - \\\midrule
   \multirow{2}{*}{Total} & \cellcolor{Gray} \textit{QA} & \cellcolor{Gray}234\,648 & \cellcolor{Gray}471\,164
   & \cellcolor{Gray}16\,095 & \cellcolor{Gray} 6\,250
   & \cellcolor{Gray} 5\,826 %(1 duplicate) 
   \\
   & \textit{Img} & 80\,831 & 120\,935 
   & 13\,645 & 1\,250
   & 2\,190
   \\
         \bottomrule
    \end{tabular}
    \caption{Summary of datasets used for evaluation of ConVQG. \textit{QA} means the number of question-answer pairs and \textit{Img} means the number of images.}
    \label{tab:datasets}
\end{table}

\subsection{Data preprocessing}\label{app:preprocessing}
The detailed data preprocessing pipeline, including dataset splitting, filtering and the creation of textual inputs, is introduced in the following paragraphs. Especially, we describe how to process different types of text inputs (such as knowledge triplets, answers, captions and fact sentences) for different datasets.

\paragraph{VQA 2.0 Small (Answer).}
Following the preprocessing method in Radial-GCN~\cite{radial2020}, we filter out question types that have ``less informative'' answers (such as ``yes/no''). Although the images for training and test are pre-assigned~\cite{karpathy2015deep}, the filtered question types of Radial-GCN are not publicly available. We try our best to make our test set quantitatively similar to previous methods (12,940 QA pairs v.s. 12,938 QA pairs). To do that, we select 28 question types out of 65 in the original annotations according to the previous method \cite{xu2018dual}.\footnote{https://github.com/yikang-li/iQAN/blob/master/data} Then we add two more question types, ``what number is'' and ``how many''. For text inputs, the answers are fed into a template: \textit{The answer to the question is [answer]}.

\paragraph{VQA 2.0 Large (Answer).}
As described in~\cite{imvqg2019}, answers in VQA 2.0 dataset are annotated with a set of 15 categories and labeled with the top 500 answers. The top 500 answers consist of 82\% of the VQA dataset, resulting in 367K training and validation examples. Because the annotations of VQA 2.0 test set are not available, following the preprocessing method in IM-VQG~\cite{imvqg2019}, we only use the training and validation set of VQA 2.0 dataset. Keeping the top 500 answers, the processed VQA 2.0 training set is split into an 80-20\% train-validation split and the processed validation set is used as the test set.

\paragraph{K-VQG (Knowledge triplet and Answer).}
For the K-VQG dataset, two types of textual constraints are used to generate questions. For knowledge triplets shown as $\textless$ subject - predicate - object$\textgreater$, we use templates to generate a short sentence based on the masked knowledge triplet. For instance, $\textless$ container - CapableOf - [MASK]$\textgreater$ is mapped to \emph{container is capable of [MASK]}. The detailed formulating method of 15 relationship categories in the paper can be found in Table \ref{tab:template}. As for the answers as text constraints, we use the same template as VQA 2.0 dataset and turn it into the sentence: \textit{The answer to the question is [answer]}.

\begin{table}[h]
\centering
\label{tab:template}
\begin{tabular}{ll}
\toprule
\textbf{Relationship}    & \textbf{Template}              \\
\midrule
UsedFor         & is used for         \\
ReceivesAction  & receives action     \\
HasA            & has a               \\
Causes          & causes              \\
HasProperty     & has a property      \\
CreatedBy       & is created by       \\
DefinedAs       & is defined as       \\
AtLocation      & is at location of   \\
HasSubEvent     & has                 \\
MadeUpOf        & is made of          \\
HasPrerequisite & has prerequisite to \\
Desires         & desires             \\
NotDesires      & not desires         \\
IsA             & is a                \\
CapableOf       & is capable of \\
\bottomrule
\end{tabular}
\caption{The template to form a sentence based on knowledge triplet.}
\end{table}

\paragraph{VQG-COCO (Caption).} We use the same split as previous work \cite{grnn2016, patro2018multimodal}, where there are 2\,500, 1\,250, and 1\,250 images for training, validation and testing. In addition, captions in the annotations are used as text constraints to give a `focus' for question generation. The dataset is different from others since there is no answer associated with questions. In this case, we use captions as textual guidance to provide some textual cues for question generation. The captions are annotated in the dataset, so they don't require any specific processing.

\paragraph{FVQA (Fact sentence).} We use the FVQA dataset as a whole for the transfer experiment, so there is no split for the dataset. In addition, FVQA dataset already has facts as textual cues, hence it doesn't require any further processing.

\section{Metrics Details}\label{app:metrics}

As briefly introduced in the main paper, we use a variety of language generation metrics to evaluate and compare ConVQG against competitors: BLEU~\cite{papineni2002bleu}, ROUGE\_L~\cite{lin2004rouge}, METEOR~\cite{denkowski2014meteor} and CIDEr~\cite{vedantam2014cider}. They assess the conformity between questions generated by a model and ground truth questions. CIDEr, a TF-IDF-based metric, is the closest to human evaluation for image description compared to the other metrics, according to~\cite{vedantam2014cider}. More details about these metrics are given below:

\begin{itemize}[leftmargin=1em, itemsep=1pt]
    \item \textbf{BLEU} (BiLingual Evaluation Understudy): it is obtained by matching text snippets with a set of reference texts. Scores are computed considering the presence of a given text segment in the reference snippets. Therefore, BLEU is a precision-based metric. Several variations of BLEU exist, depending on the number of $n$-grams to match in the reference text (BLEU-1, BLEU-2, $\ldots$, BLEU-$n$). BLEU-1 considers only 1-grams, while BLEU-$n$ considers $k$-grams with $k$ varying from 1 to $n$.
    \item \textbf{ROUGE\_L} (Recall-Oriented Understudy for Gisting Evaluation): it gathers several metrics to evaluate the generated text against the reference. Contrary to BLEU, these metrics are recall-based. In particular, we used the ROUGE\_L variant in this work, which measures the longest common sub-sequence between the generated sequence and the reference.
    \item \textbf{METEOR} (Metric for Evaluation of Translation with Explicit ORdering): it is classically used for machine translation evaluation. METEOR is based on the harmonic mean of 1-gram precision and recall, where recall weighs more than precision. It uses exact word matching and the ability to stem and match synonyms.
    \item \textbf{CIDEr} (Consensus-based Image Description Evaluation): it was conceived to evaluate the correspondence between the generated text and the reference, especially for image descriptions. After stemming and representing every text snippet as a set of 1 to 4 grams, CIDEr is computed by first calculating the co-occurrences of these $n$-grams with reference $n$-grams. Then, the cosine similarity between $n$-grams of the generated text and the references is computed, giving less weight to frequent $n$-grams (which are likely to be less informative). 
\end{itemize}

\section{Experimental Setting Details}
Here we give more details about the hyper-parameter settings, mainly about the hyper-parameters in the text decoder and training. For the image input, the image size is set to 480. For the BERT model, the number of hidden layers is 12 and the number of attention heads is 12. For beam search decoding during inference, the number of beams is set to 3. For training, the initial learning rate is 2e-5 and weight decay is set to 0.05.

For more details about the experimental environment, we used torch 1.11.0+cu113 and torchvision 0.12.0+cu113. GPU details are shown in the paper.

\begin{table}[h]
\centering

\label{tab:parameter_supp}
\begin{tabular}{ll}
\toprule
\textbf{Parameter}    & \textbf{Value}              \\
\midrule
initial learning rate         & 2e-5         \\
image size  & 480     \\
weight decay            & 0.05               \\
number of beams          & 3              \\
number of attention heads     & 12      \\
number hidden layers       & 12      \\
\bottomrule
\end{tabular}
\caption{The template to form a sentence based on knowledge triplet.}
\end{table}

\section{Quantitative Results}

\subsection{Transfer results on FVQA dataset}

Besides the standard visual question generation settings, our model can generate questions for open-domain images and texts using the inference mode. To demonstrate the generalization ability of the proposed ConVQG model, we train it on the K-VQG dataset and test its performance on the FVQA dataset. There were some possible overlaps over images in the K-VQG and FVQA datasets (images from the COCO validation dataset), but the text inputs are annotated differently. More specifically, the text input of each image in the FVQA dataset is a fact sentence rather than a knowledge triplet.

\begin{table}[h]
\centering
  \setlength{\tabcolsep}{6pt}
  \small{
    \begin{tabular}{rcccc}
    \toprule
      & \multicolumn{1}{p{3.8em}}{BLEU-4} & \multicolumn{1}{p{3.8em}}{METEOR} & \multicolumn{1}{p{3.8em}}{ROUGE\_L  } & \multicolumn{1}{p{3.8em}}{CIDEr} \\
\midrule
           \multicolumn{1}{p{3.5em}}{\textbf{ConVQG}$_{B}$} & \multicolumn{1}{c}{2.96 } & \multicolumn{1}{c}{\textbf{13.78 }} & \multicolumn{1}{c}{23.67 } & \multicolumn{1}{c}{0.37 } \\

           \multicolumn{1}{p{3.5em}}{\textbf{ConVQG}$_{IT}$} & \multicolumn{1}{c}{\textbf{3.04 }} & \multicolumn{1}{c}{13.77 } & \multicolumn{1}{c}{\textbf{23.68 }} & \multicolumn{1}{c}{\textbf{0.41 }} \\
\bottomrule
    \end{tabular}%
    }
    \caption{Transfer results on FVQA dataset. Both the baseline method \textbf{ConVQG}$_{B}$ and the proposed \textbf{ConVQG}$_{IT}$ are trained on the K-VQG dataset with knowledge triplets as text input. We report the evaluation results on the whole FVQA dataset.}
  \label{tab:fvqa}%
\end{table}%

\begin{table*}[t]
\centering
\vspace{-0.5em}
\setlength{\tabcolsep}{0pt}
\renewcommand{\arraystretch}{0.8}

\small
\begin{tabular}{m{0.3\linewidth}>{\centering \arraybackslash}m{0.15\linewidth}>{\centering \arraybackslash}m{0.15\linewidth}>{\centering \arraybackslash}m{0.15\linewidth}>{\centering \arraybackslash}m{0.15\linewidth}}
\toprule
\emph{\footnotesize Method} & {\footnotesize BLEU-1} & {\footnotesize METEOR} & {\footnotesize ROUGE\_L} & {\footnotesize CIDEr} \\
\midrule
I2Q \cite{grnn2016} & 19.2 & 19.7 & - & - \\
Creative \cite{jain2017creativity} & 35.6 & 19.9 & - & - \\
MDN \cite{patro2018multimodal} & 36.0 & 23.4  & 41.8 & 0.51 \\
MC-BMN \cite{patro2020deep} & 40.7 & 22.6  & \textbf{41.9} & 0.50 \\
{\textbf{ConVQG$_{IT}$}} & \textbf{50.2} & \textbf{26.4} & 40.3 & \textbf{0.56} \\
\bottomrule
\end{tabular}%
\caption{Results on the VQG-COCO test sets.}
\label{tab:vqgcoco_appendix}%
\vspace{-0.5em}
\end{table*}

\begin{table*}[t]
  \centering
  \setlength{\tabcolsep}{2pt}

    \begin{tabular}{>{\centering \arraybackslash}m{0.17\linewidth}m{0.40\linewidth}>{\centering \arraybackslash}m{0.09\linewidth}>{\centering \arraybackslash}m{0.09\linewidth}>{\centering \arraybackslash}m{0.09\linewidth}}
    \toprule
          \emph{Text constraint} & \emph{Method} & {BLEU-4} & {METEOR} & {CIDEr} \\
    \midrule
    \multirow{2}{*}{Answer}
   &{IM-VQG \cite{imvqg2019}} & 12.37  & 16.65  & 0.39  \\
    &\textbf{ConVQG}$_{IT}$ & \textbf{14.30} & \textbf{18.67}  & \textbf{0.78} \\
    
    \midrule
    \multirow{2}{=}{\centering Knowledge Triplet}
    &{K-VQG \cite{kvqg2023}} & 18.84  & \textbf{22.79}  & 1.31  \\
    &\textbf{ConVQG}$_{IT}$ & \textbf{20.01} & 22.66  & \textbf{1.53} \\
    \bottomrule
    \end{tabular}%
      \caption{Results on K-VQG dataset.}
  \label{tab:kvqg_appendix}%
\end{table*}%

\begin{table*}[t]
  \centering
  \setlength{\tabcolsep}{4pt}
  
    \begin{tabular}{>{\centering \arraybackslash}m{0.1\linewidth}m{0.25\linewidth}>{\centering \arraybackslash}m{0.09\linewidth}>{\centering \arraybackslash}m{0.09\linewidth}>{\centering \arraybackslash}m{0.09\linewidth}>{\centering \arraybackslash}m{0.09\linewidth}>{\centering \arraybackslash}m{0.09\linewidth}}
    \toprule
        \emph{Test set}   &  \emph{Method}     & {BLEU-1} & {BLEU-4} & {METEOR} & {ROUGE\_L} & {CIDEr}   \\
          \midrule
        {\multirow{9}{*}{Small}} & {SAT \cite{xu2015show}} & {49.4} & {23.1} & {24.4} & {53.4} & {1.65 } \\
         & {DL-VQG \cite{xu2018dual}} & {50.7} & {24.4} & {26.4} & {55.9} & {1.88 } \\
           & {IVQA \cite{liu2018inverse}} & {50.2} & {23.9} & \textbf{35.7} & {55.3} & {1.84 }  \\
          & {IM-VQG \cite{imvqg2019}} & {51.3} & {24.8} & {26.3} & {56.3} & {1.94 }  \\
          & {iQAN \cite{li2018visual}} & {52.6} & {27.1} & {26.8} & {56.9} & {2.09 }  \\
           & {Radial-GCN \cite{radial2020}} & {53.4} & {27.9} & {27.1} & {57.2} & {2.10 } \\
          & {MOAG \cite{moag2021}} & {58.8} & {28.1} & {27.8} & {60.4} & {2.39 } \\
           & {\textbf{ConVQG$_{IT}$}} & {\textbf{59.9}} & {\textbf{33.1}} & {30.0} & {\textbf{62.6}} & {\textbf{2.79}}  \\
          \midrule
          {\multirow{3}[2]{*}{Large}} & {C3VQG \cite{2021c3vqg}} & {41.9 } & {10.0 } & {13.6 } & {42.3 } & {0.47 }  \\
          & {IM-VQG \cite{imvqg2019}} & {\textbf{50.1} } & {16.3 } & {20.6 } & {39.6 } & {0.94 } \\
           
           & {\textbf{ConVQG$_{IT}$}} & {45.8} & {\textbf{22.4}} & {\textbf{21.8}} & {\textbf{47.4}} & {\textbf{1.78}}  \\
\bottomrule
    \end{tabular}%
    \caption{Results on VQA 2.0 dataset small/large test set.}
  \label{tab:vqa2_appendix}%
\end{table*}%

Experimental results can be found in Table \ref{tab:fvqa}, where the proposed contrastive ConVQG$_{IT}$ model is compared with the non-contrastive baseline model ConVQG$_{B}$ in a transfer setting. The contrastive method gains slight improvements on all metrics except METEOR (0.08\% on BLEU-4, 0.01\% on ROUGE\_L and 0.04\% on CIDEr).

\subsection{Comparison method details} \label{app:baselines}

This section reports additional results of VQG models from the literature. Tables~\ref{tab:vqgcoco_appendix}, ~\ref{tab:kvqg_appendix} and~\ref{tab:vqa2_appendix} present the complete list of results in the VQG-COCO, the K-VQG and the VQA 2.0 datasets, respectively. The comparison method details are as follows.

\begin{itemize}[itemsep=1pt, leftmargin=1em]
    \item I2Q~\cite{grnn2016} only uses the image to generate the questions.
    \item K-VQG~\cite{kvqg2023} jointly encodes the image and the target knowledge (treated as a sequence of words) using a pre-trained UNITER encoder \cite{chen2020uniter}, followed by an autoregressive text decoder to generate the question.
\end{itemize}

\begin{itemize}[itemsep=1pt, leftmargin=1em]
    \item SAT~\cite{xu2015show} (``Show, Attend and Tell'') is one of the earliest works incorporating soft and hard attention into image analysis. This model is built to generate captions, with a CNN as image encoder and an LSTM as decoder.
        \item  DL-VQG~\cite{xu2018dual} (``Dual Learning for Visual Question Generation'') uses reinforcement learning to jointly perform VQA and VQG.
            \item IVQA~\cite{liu2018inverse} implements a conditional question generation model to make use of the answer to generate the question.
\item iQAN~\cite{li2018visual} is similar to DL-VQG. Same as IVQA, it takes the answers as inputs to help generating the questions.
    \item IM-VQG~\cite{imvqg2019} (`` Information Maximizing Visual Question Generation'') uses both the answer and its category to condition the question generation, maximizing the mutual information of the image, the question and the answer. When the dataset has no category, the answer itself is considered as one. 
    
    \item Radial-GCN~\cite{radial2020} uses a radial Graph Convolutional Network (GCN) to represent the image content and matches the core information for question generation.
    \item MOAG~\cite{moag2021} (``Multiple Objects-Aware Visual Question Generation'') is the SOTA method on VQA 2.0, proposing to use answers about multiple objects to generate questions.

    \item C3VQG~\cite{2021c3vqg} uses VAE to exploit the visual information for question generation without groundtruth answers.

    \item Creative~\cite{jain2017creativity} combines variational autoencoders with long short-term memory networks to generate creative questions.

    \item MDN~\cite{patro2018multimodal} (Multimodal Differential Network) is a multimodal network that uses exemplars for obtaining the relevant context to produce natural and engaging questions by triplet losses.

    \item MC-BMN~\cite{patro2020deep} is a deep Bayesian learning model for probabilistic question generation based on multimodal cues.
\end{itemize}

\section{Qualitative Results}

\begin{figure}[t]
	\centering
	\includegraphics[width=0.999\linewidth]{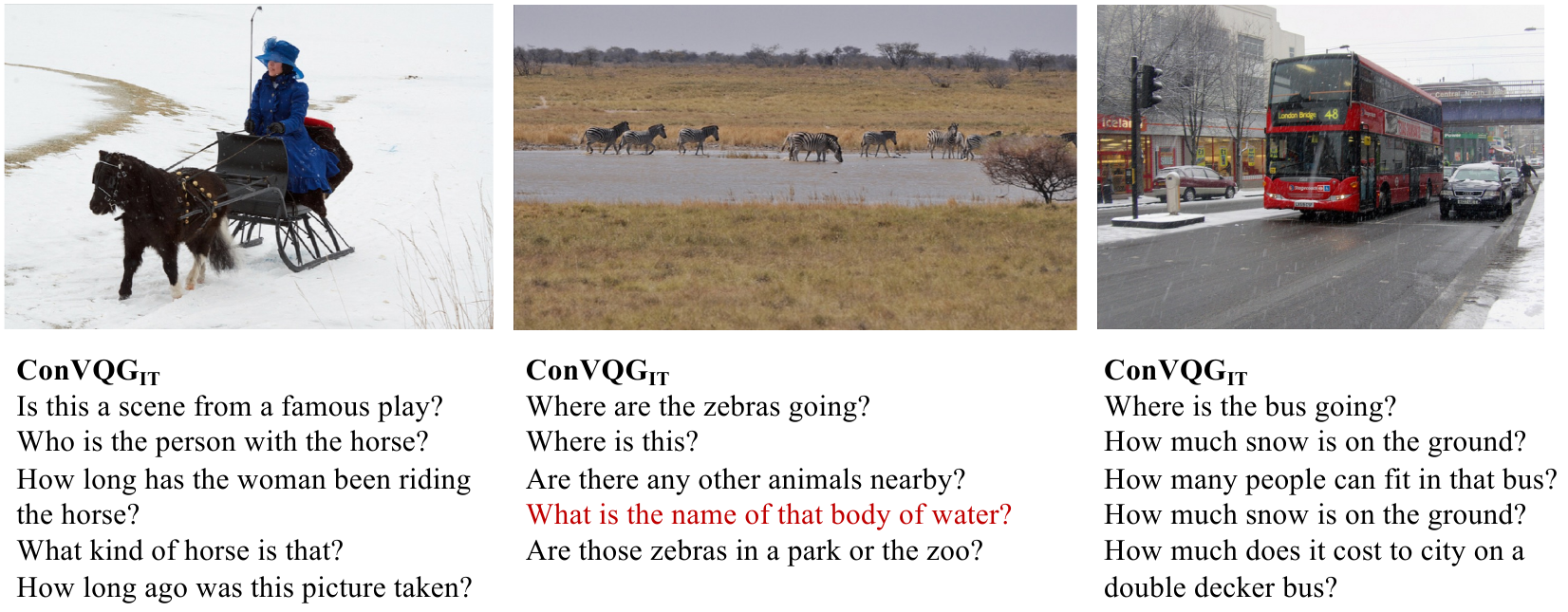}
	\vspace{2mm}
	\caption{Examples from the VQG-COCO dataset. Since we take captions are constraints in this dataset, which gives more flexibility to the question generation system, the generated questions are more diverse. \textcolor{myred}{Red color} indicates wrong expressions, not related to the image.}
	\label{fig:exvqgcoco}
\end{figure}

\textbf{Diversity.} Examples from the VQG-COCO dataset are shown in Fig.~\ref{fig:exvqgcoco}. Since there is not necessarily an answer associated with the question, captions are used as text inputs. On one hand, it is more difficult to use captions to guide the question generation, since captions are usually the description of the whole image. On the other hand, the uncertainty also brings the diversity of question content. Without obvious guidance for questions, the questions can be anything that is related to image content (captions). The results show that in this case, questions generated by ConVQG can be more natural, creative and diverse. We take them as a special case for ConVQG applications.

\begin{figure}[h]
	\centering
	\includegraphics[width=0.99\linewidth]{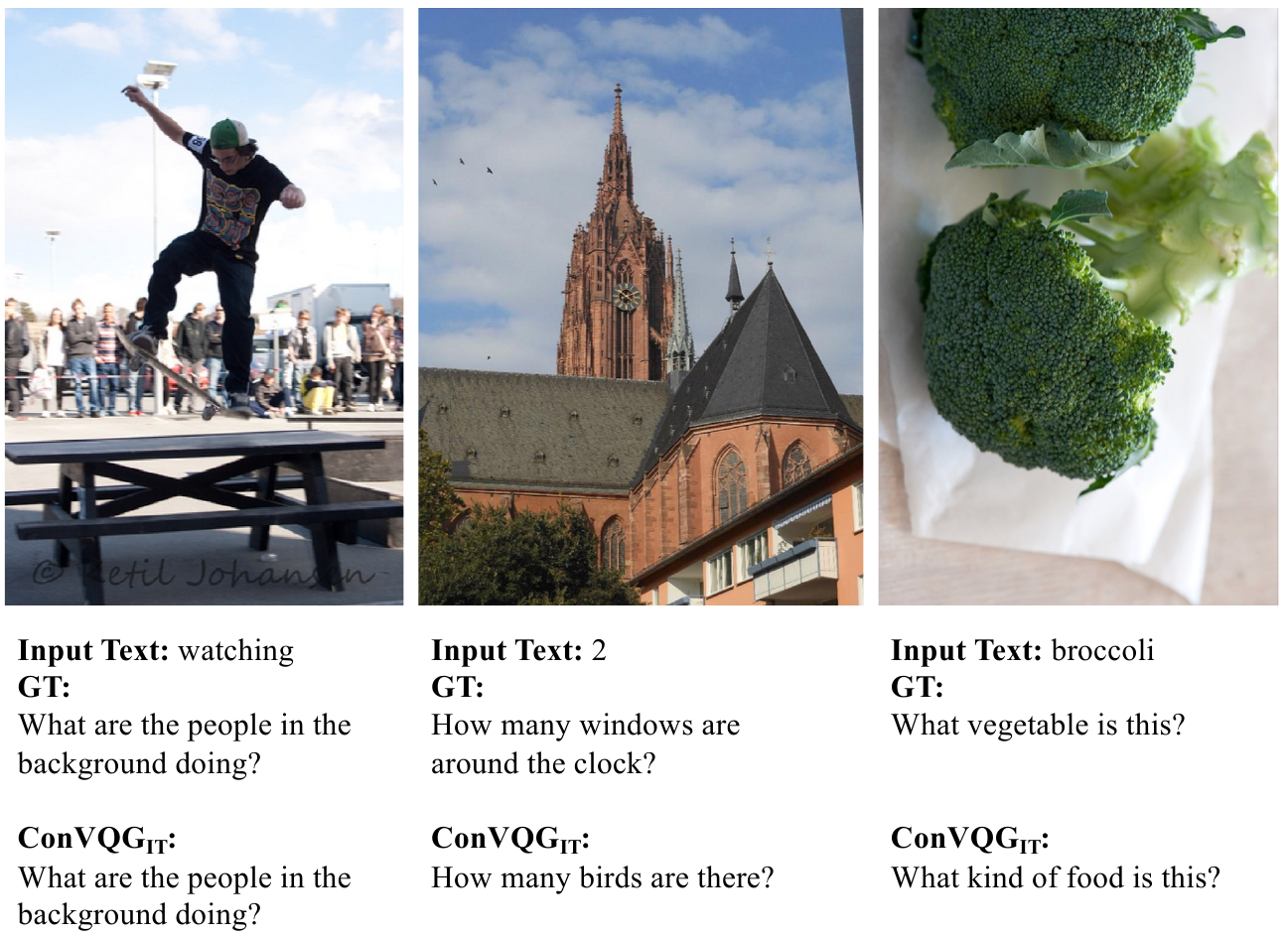}
	\caption{Examples from the VQA 2.0 small test set. The answers are used as text inputs.}
	\label{fig:exvqa}
\end{figure}

\begin{figure*}[h]
	\centering
	\includegraphics[width=0.9\linewidth]{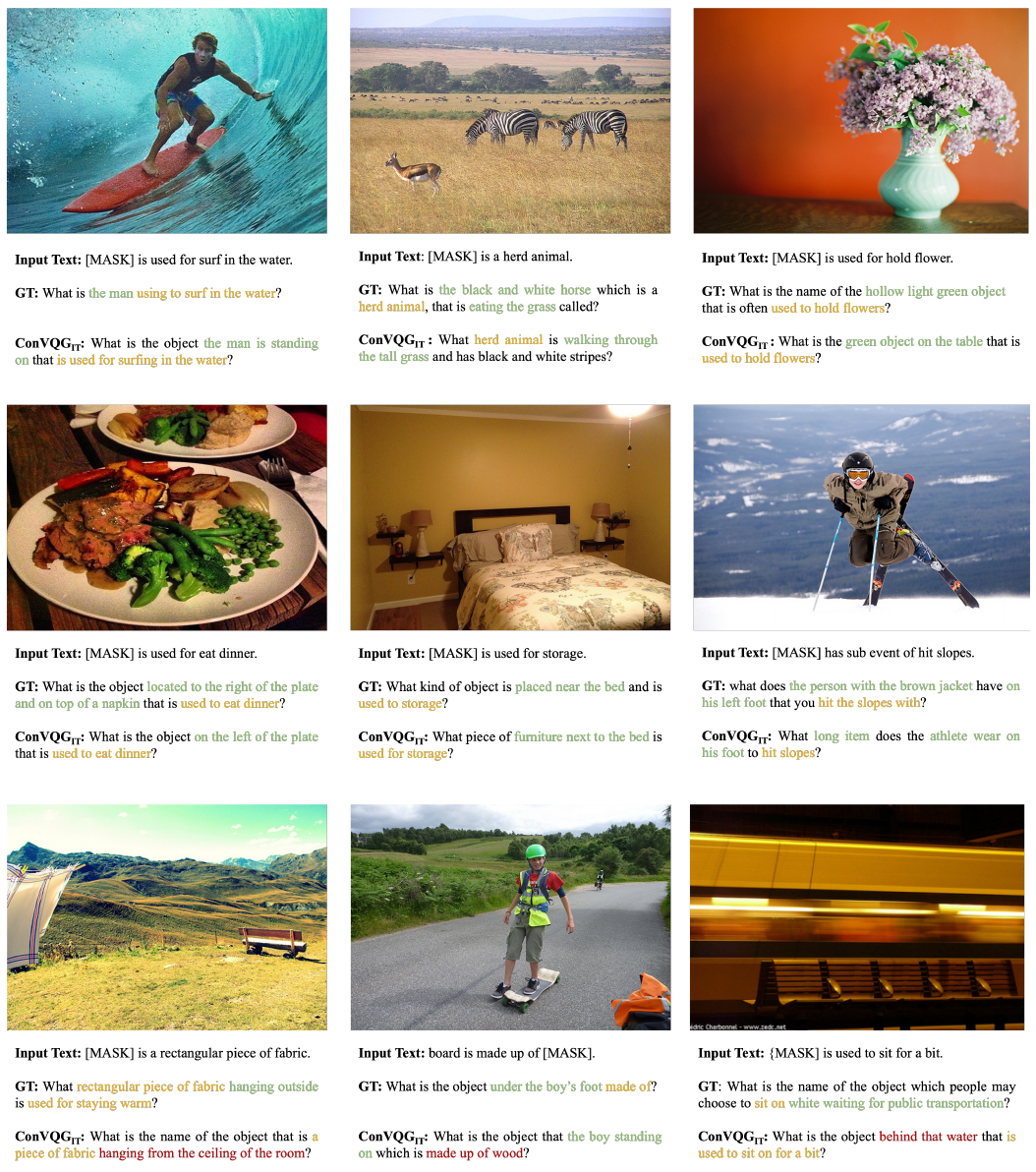}
	\caption{Additional examples from K-VQG dataset. The first and second rows show examples in which the generated questions are successfully grounded to both image and text. The last row shows some failure cases where the model provides wrong information about image content or text constraints. In the text, \textcolor{mylime}{green color} denotes the sequence that is related to image content, while \textcolor{myyellow}{yellow color} denotes the information that is carried by the text input. \textcolor{myred}{Red color} indicates wrong expressions, not related to the image or the text input.}
	\vspace{2mm}
	\label{fig:examples}
\end{figure*}

\paragraph{Different text inputs.} We also show examples from the VQA 2.0 dataset as well as more examples from the K-VQG dataset in Fig.~\ref{fig:exvqa} and Fig.~\ref{fig:examples}, respectively. For the VQA 2.0 dataset, the model takes answers as text inputs, while for the K-VQG dataset, text constraints can be answers or knowledge triplets. Comparing those two figures we can see, different text inputs lead to different types of questions. Answers are more precise guidance, where the model can `guess' the question types from the answers sometimes. For example, if the answer is `green' then the question probably is about the color of an object in the image. On the other hand, knowledge triplets give external commonsense knowledge that is difficult to obtain from the image directly. By providing this, questions are more informative, diverse and challenging.

\paragraph{Error analysis.}
We also provide more examples from the K-VQG dataset, especially some failure cases in Fig.~\ref{fig:examples}. The first two rows show more examples where the generated questions from the proposed ConVQG method can be both image-grounded and text-guided. The last row presents some of the failure cases. For the first and third examples of failure cases (Columns 1 and 3, Row 3), the model generates a question with respect to the text input but adds inappropriate descriptions of image content (e.g. \emph{the ceiling of the room} and \emph{behind the water}). For the first example, the model selects the most likely place where the \emph{fabric} will appear but doesn't pay attention to the image content. For the third example, the model incorrectly detects \emph{water} from the image. For the second failure case (Column 2, Row 3), the model fails to constrain the question by the input text \emph{board is made up of something}, on the contrary, it generates the questions based on the most likely answer \emph{wood}.

\section{Human Evaluation}\label{app:human_eval}

We use MTurk to get human preference in order to evaluate the effect of the contrastive branch of ConVQG.

\begin{figure*}[h]
	\centering
	\includegraphics[width=0.9\linewidth]{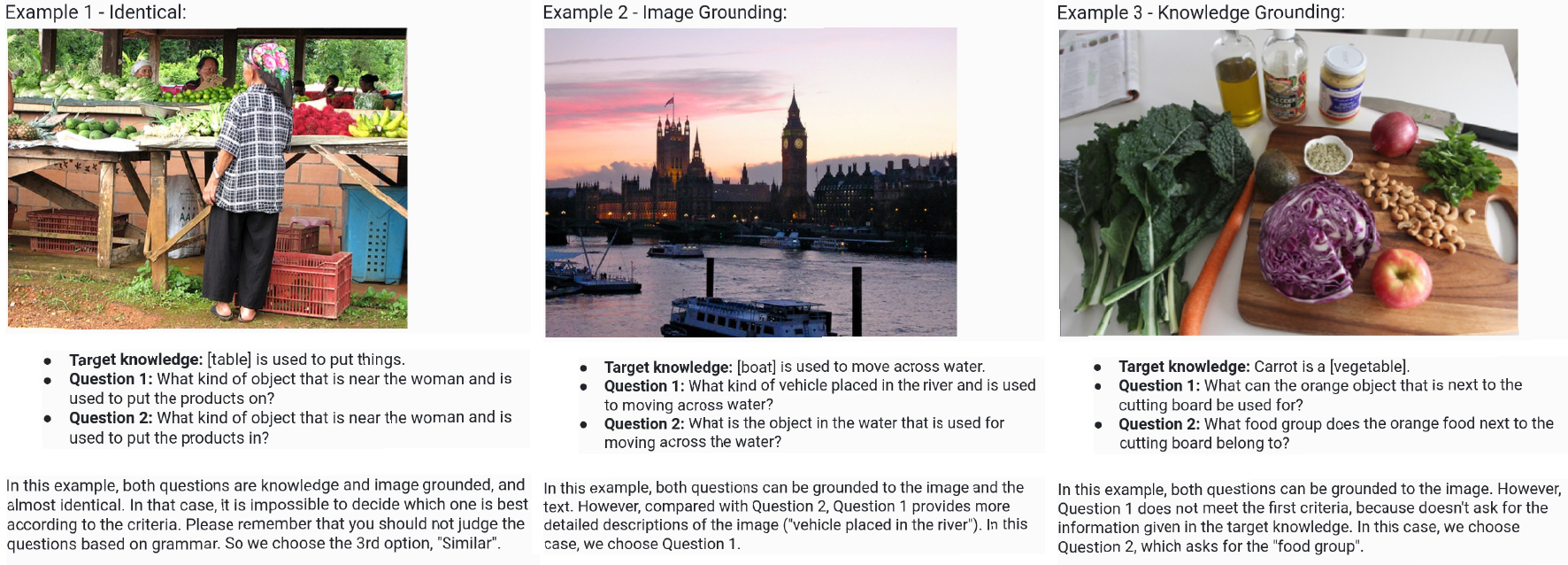}
	\caption{Examples given as instructions for MTurk annotators. We give three different examples: identical, image grounding and knowledge grounding.}
	\label{fig:Mturk_examples}
\end{figure*}

\paragraph{Selection of examples to evaluate.} 
We asked workers to evaluate 500 examples of the test set of K-VQG dataset, comparing ConVQG$_{B}$ and ConVQG$_{IT}$ generated questions.
From the set of 3207 examples in the test set of K-VQG, we deduplicated images and knowledge triplets. We also removed cases where the baseline model ConVQG$_{B}$ and the contrastive model ConVQG$_{IT}$ output the exact same questions (155 cases, 4.8\% of the test set). Then, we sampled 500 examples, randomly swapping the two questions to avoid bias in the comparison.
On top of the two questions to compare and the image, we provide the workers with the knowledge triplet containing the answer to the question; moreover, we highlight in the sentence which section corresponds to the answer, as seen in the examples given to the workers in Fig.~\ref{fig:Mturk_examples}.

\begin{figure*}[h]
	\centering
        \includegraphics[width=0.45\linewidth]{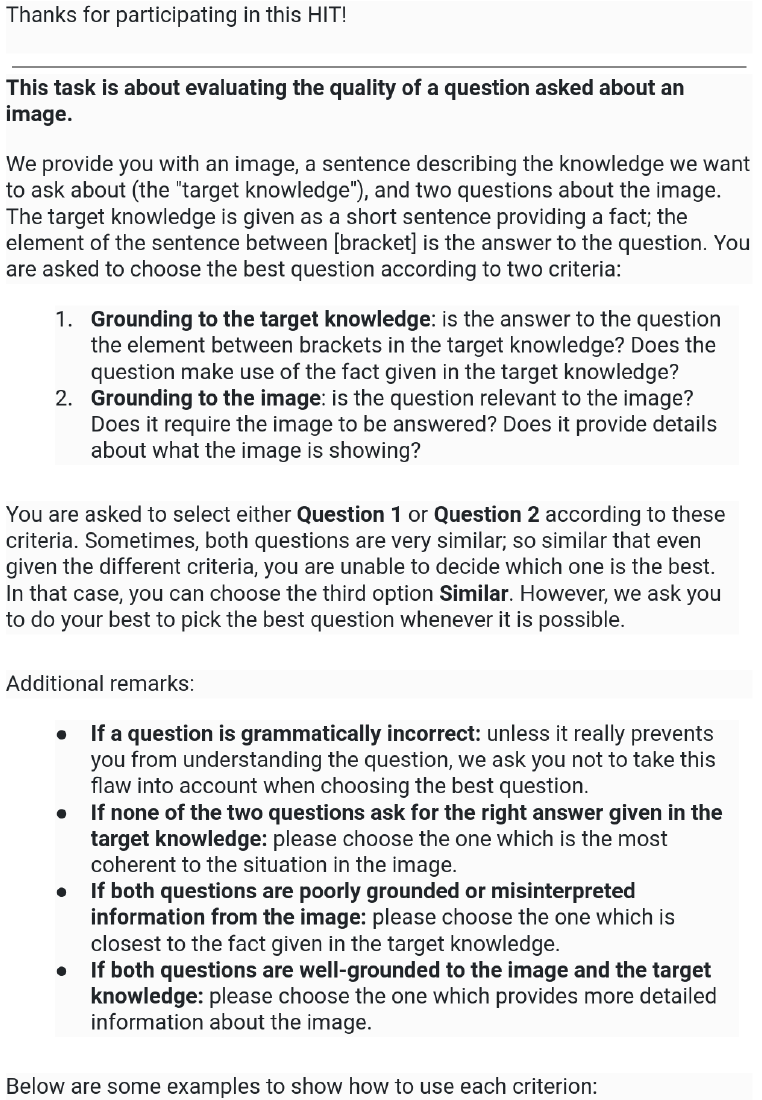}
	\caption{Instructions given to crowd workers on MTurk.}
	\label{fig:mturk_instructions}
\end{figure*}

\paragraph{Instructions given to crowd workers.} 
On top of the examples in Fig. \ref{fig:Mturk_examples}, we gave detailed instructions to the workers; they can be found in Fig. \ref{fig:mturk_instructions}. We list criteria to focus on when selecting the best question relative to the image and the knowledge triplet (which we call \textit{target knowledge} in the instructions). The two main criteria are the grounding of the question to the image and to the knowledge triplet. We specifically asked the workers not to focus on the grammatical correctness of the question to make their choice. Indeed, the difference in architecture and training of the two models we compare should not lead to a significant variation in their ability to generate grammatically correct text; hence, we want the workers to focus on the grounding aspect of the questions. Workers are given the possibility to choose none of the two questions if they consider that the similarity between them is too high to make a meaningful choice. After removing examples where the two questions are identical, however many examples remain where only a few words differ between the two questions. Each worker was given 5 examples per hit. Each hit was only seen by one worker. The workers were pre-selected according to their performance on other tasks.

\paragraph{Overall results.} The overall results are shown in Table~\ref{tab:app_mturk_res}, where ConVQG$_{IT}$ gets 55 more votes than ConVQG$_{B}$ among 500 samples.

\begin{table}[t]
    \centering
    
    \begin{tabular}{lc}
\toprule
\emph{Method} & \emph{Votes}  \\
\midrule
  \textbf{ConVQG}$_{IT}$  &          236 \\
 \textbf{ConVQG}$_{B}$ &           183 \\
  Similar &           81 \\
\bottomrule
\end{tabular}
\label{tab:app_mturk_res}
\caption{Results from MTurk. The vote means the number of times chosen by the annotator in pairwise comparison.}
\end{table}

\end{document}